\RequirePackage{fix-cm}
\documentclass[twocolumn]{svjour3}          

\smartqed  
\usepackage{times}
\usepackage{epsfig}
\usepackage{graphicx}
\usepackage{subcaption}
\usepackage{amsmath}
\usepackage{amssymb}
\usepackage{bm}
\usepackage{textcomp}
\usepackage{enumitem}
\usepackage{xcolor}
\usepackage[pagebackref=true,breaklinks=true,letterpaper=true,colorlinks,bookmarks=false]{hyperref}
\usepackage[numbers]{natbib}
\newcommand{\eg}{\text{e.g.}}
\newcommand{\ie}{\text{i.e.}}
\newcommand{\re}[1]{{\color{black}{#1}}}

\begin{document}
\title{Learning to Collocate Visual-Linguistic Neural Modules for Image Captioning
}


\author{Xu Yang, Hanwang Zhang, Chongyang Gao, Jianfei Cai\\
}


\institute{Xu Yang \at
              School of Computer Science and Engineering, Southeast University, China, \\
              \email{101013120@seu.edu.cn}           
          \and
          Hanwang Zhang \at
              School of Computer Science and Engineering, Nanyang Technological University, Singapore, \\
              \email{hanwangzhang@ntu.edu.sg} \\
              He is the corresponding author.
        \and
        Chongyang Gao \at
              Computer Science Department, Northwestern University, United States,\\
              \email{cygao@u.northwestern.edu}
        \and
        Jianfei Cai\at
              Faculty of Information Technology, Monash University, Australia, \\
              \email{Jianfei.Cai@monash.edu}  
}

\date{Received: date / Accepted: date}

\maketitle
\begin{abstract}
\re{Humans tend to decompose a sentence into different parts like \textsc{sth do sth at someplace} and then fill each part with certain content.} Inspired by this, we follow the \textit{principle of modular design} to propose a novel image captioner: learning to Collocate Visual-Linguistic Neural Modules (CVLNM). Unlike the \re{widely used} neural module networks in VQA, where the language (\ie, question) is fully observable, \re{the task of collocating visual-linguistic modules is more challenging.} This is because the language is only partially observable, for which we need to dynamically collocate the modules during the process of image captioning. To sum up, we make the following technical contributions to design and train our CVLNM: 1) \textit{distinguishable module design} --- \re{four modules in the encoder} including one linguistic module for function words and three visual modules for different content words (\ie, noun, adjective, and verb) and another linguistic one in the decoder for commonsense reasoning, 2) a self-attention based \textit{module controller} for robustifying the visual reasoning, 3) a part-of-speech based \textit{syntax loss} imposed on the module controller for further regularizing the training of our CVLNM. Extensive experiments on the MS-COCO dataset show that our CVLNM is more effective, \eg, achieving a new state-of-the-art 129.5 CIDEr-D, and more robust, \eg, being less likely to overfit to dataset bias and suffering less when fewer training samples are available. Codes are available at \url{https://github.com/GCYZSL/CVLMN}. 
\end{abstract}

\section{Introduction}
\label{sec:intro}

\begin{figure}[t]
\centering
    \begin{subfigure}[t]{.9\linewidth}
        \includegraphics[width=1\linewidth,trim = 5mm 5mm 5mm 5mm,clip]{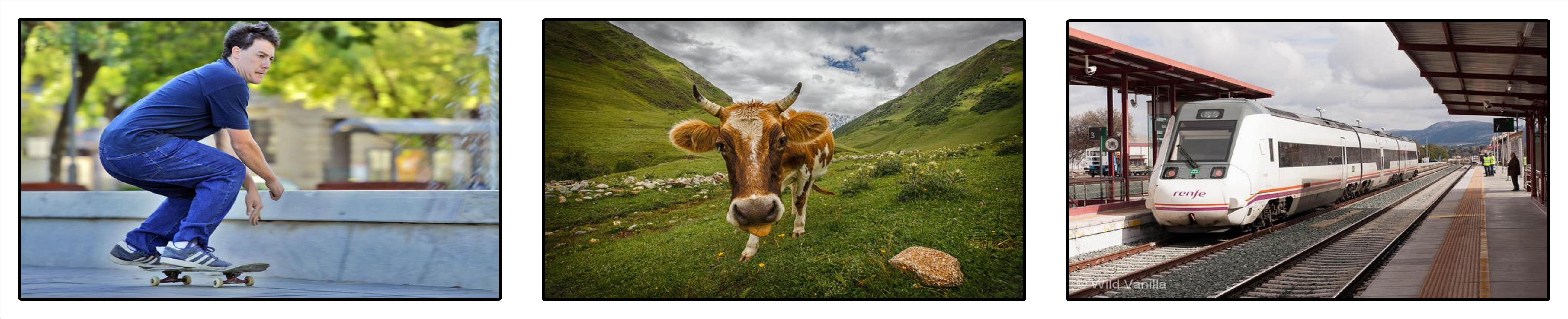}
        \caption{Three diverse images. }
        \label{fig:introa}
    \end{subfigure}
    \begin{subfigure}[t]{.9\linewidth}
        \includegraphics[width=1\linewidth,trim = 5mm 5mm 5mm 5mm,clip]{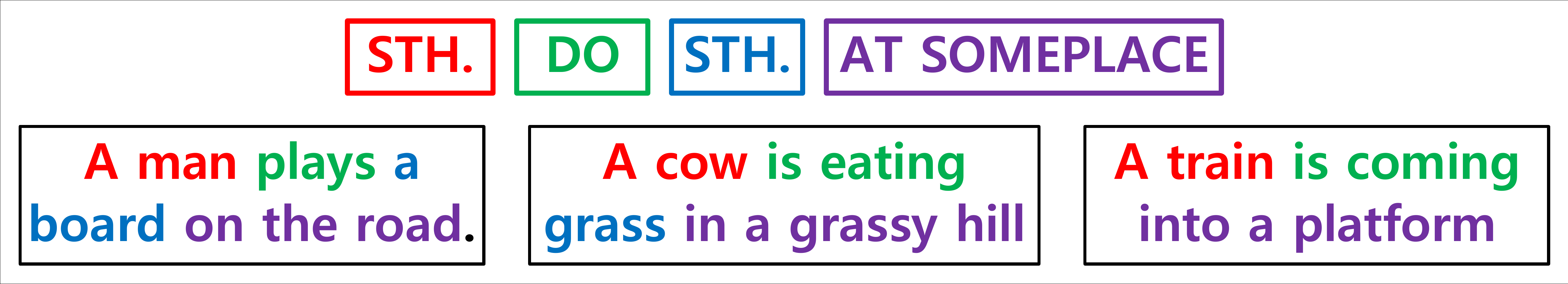}
        \caption{Three captions with the same sentence pattern.}
        \label{fig:introb}
    \end{subfigure}
    \begin{subfigure}[t]{.9\linewidth}
        \includegraphics[width=1\linewidth,trim = 5mm 5mm 5mm 5mm,clip]{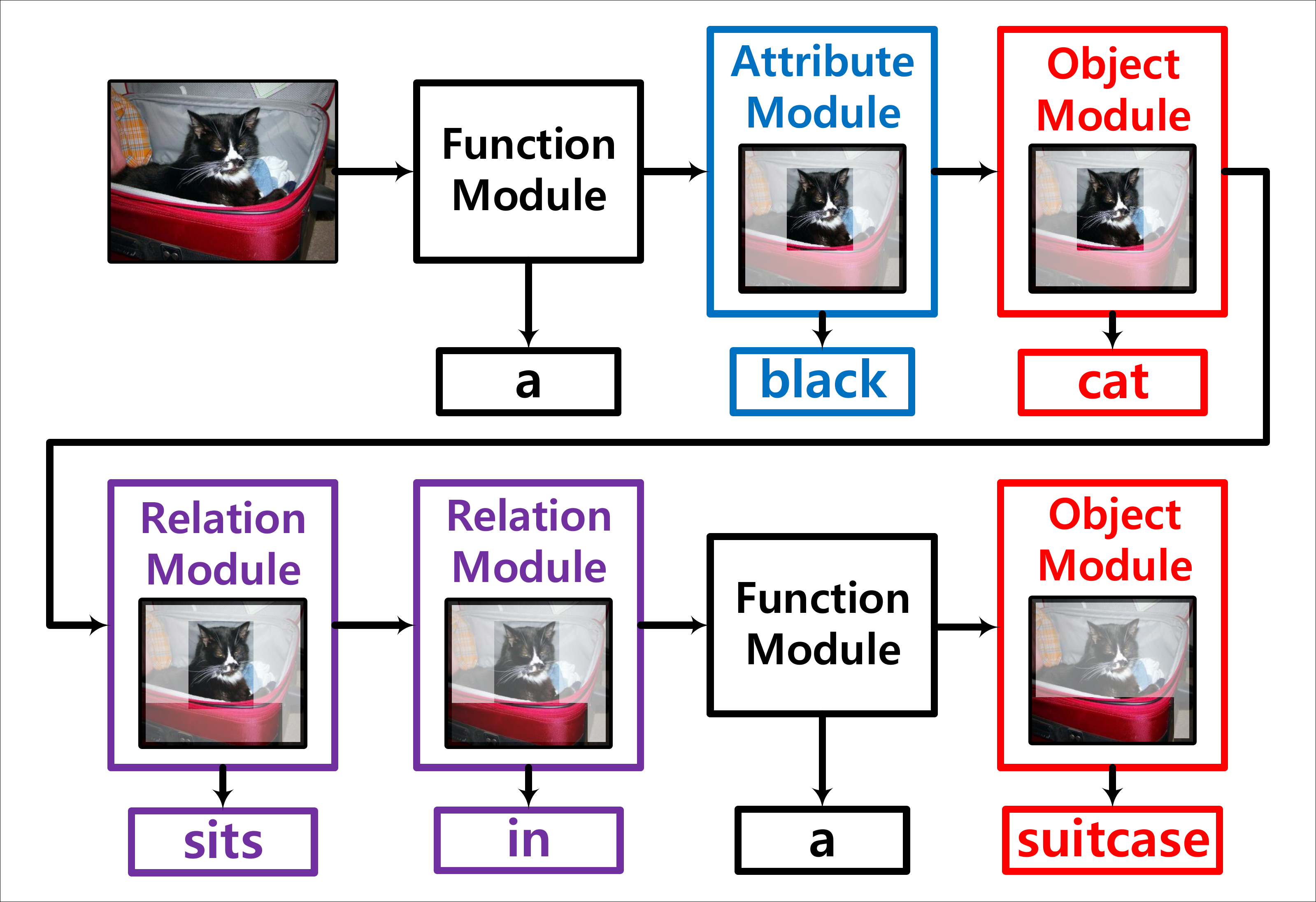}
        \caption{The captioning process of CVLNM.}
        \label{fig:introc}
    \end{subfigure}
     \caption{\re{Motivated by the the principle of the modular design, we Collocate Visual-Linguistic Neural Modules (CVLNM) for image captioning.}}
\end{figure}

\begin{figure*}[t]
\centering
\includegraphics[width=1\linewidth,trim = 5mm 5mm 5mm 5mm,clip]{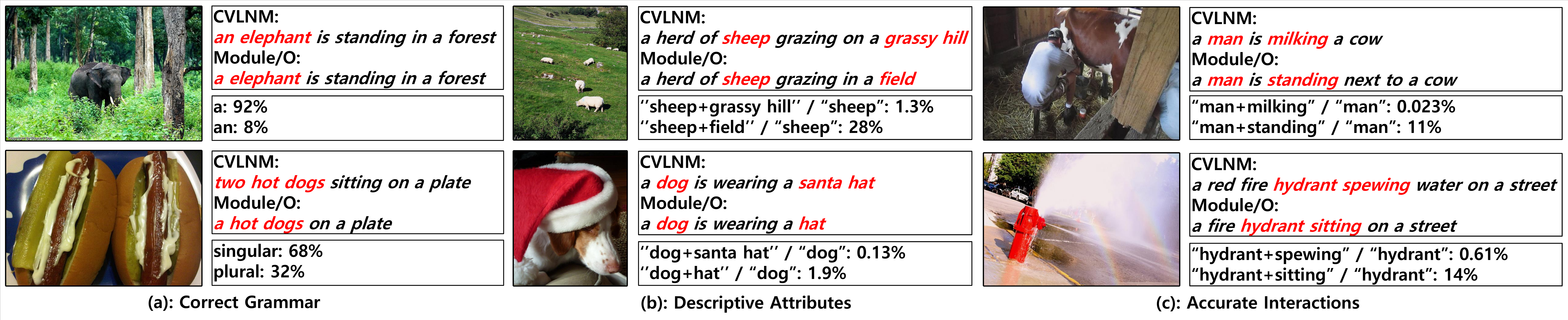}
  \caption{By comparing our CVLNM with the baseline Module/O, which only uses \textsc{object} module in the encoder (an upgraded version of Up-Down~\cite{anderson2018bottom}), we have three interesting findings in tackling dataset bias: (a) more accurate grammar, where \% in the bottom box for each image denotes the frequency of a certain pattern in MS-COCO, (b) more descriptive attributes, and (c) more accurate object interactions. The ratio ./. denotes the percentage of co-occurrence, \eg, ``sheep+field''/``sheep'' = 28\% means that ``sheep'' and ``field'' contributes the 28\% occurrences of ``sheep''. We can see that CVLNM outperforms Module/O even with highly biased training samples.}
\label{fig:intro2}
\end{figure*}

\re{Let us} describe the three images in Fig.~\ref{fig:introa}. Most people will compose sentences varying vastly from image to image. In fact, the ability to use diverse language to describe the colorful visual world is a gift to humans, but a formidable challenge to machines. Although recent advances in visual representation learning~\cite{he2016deep,ren2015faster} and language modeling~\cite{hochreiter1997long,vaswani2017attention} demonstrate \re{an impressive power} of modeling the diversity in their respective modalities, it is still far from establishing a robust cross-modal connection between them. 

Actually, most visual reasoning systems tend to overfit to dataset biases and thus fail to reproduce the diversity of our world --- the more complex the task is, the severer the overfitting will be, such as VQA~\cite{johnson2017clevr,shi2019explainable,tang2019learning}, image paragraph generation~\cite{krause2017hierarchical}, scene graph generation~\cite{chen2019counterfactual}, and visual dialog~\cite{das2017visual,Niu_2019_CVPR}. Similarly, modern image captioning systems also easily exploit dataset bias to generate the captions and sometimes even without looking at the image~\cite{rohrbach2018object}.
For example, in MS-COCO~\cite{lin2014microsoft} captioning training set, as the co-occurrence frequency of ``man'' and ``standing'' is 11\%  (Fig.~\ref{fig:intro2} (c)), a state-of-the-art captioner~\cite{anderson2018bottom} is very likely to generate ``man standing'', regardless of their actual relationship is ``milking'', whose co-occurrence frequency is only 0.023\%.

Unfortunately, unlike a visual concept in ImageNet which has 650 training images on average~\cite{deng2009imagenet} for image classification, a specific sentence in MS-COCO only corresponds to \emph{one single} image~\cite{lin2014microsoft}, which is extremely scarce in the conventional view of supervised training. 
\re{However, humans generally do not require many training samples to perform captioning. Since we can decompose a sentence into different parts and then fill suitable contents into each part, \eg, as shown in Figure~\ref{fig:introb}. 
}
Although substantial progress has been made in the past few years since \re{ since the early work of Vinyals \textit{et al.}~\cite{vinyals2015show}}, such crucial technique has not been well-studied in the field of image captioning.

The missing crux is the \textit{principle of modular design}, which makes a system less likely to overfit to dataset bias since each module only needs to learn its related supervision instead of being entangled by unrelated knowledge~\cite{Marr:1982:VCI:1095712,bengio2013representation}. Apart from confronting dataset bias, modular design also enriches the flexibility and extensibility of a learning system that researchers can insert suitable modules for achieving specific goals~\cite{andreas2016neural,hu2017learning}.


Motivated by these advantages, we follow the modular design principle~\cite{Marr:1982:VCI:1095712,bengio2013representation} and propose learning to Collocate Visual-Linguistic Neural Modules (CVLNM), which decomposes captioning into a series of sub-tasks that each one is solved by a specific module. To cover the frequently appearing part-of-speech of a sentence, we design an empirically complete set of modules in the encoder, including \textsc{object} module for nouns, \textsc{attribute} module for adjectives, \textsc{relation} module for verbs, prepositions, and quantifiers, and \textsc{function} module for other function words (see Section~\ref{subsec:modules}). Among them, the former three are visual modules and the last one is a linguistic module. We apply different inductive biases to design these modules and thus they can learn distinguishable information~\cite{locatello2018challenging} to solve specific sub-tasks, \ie, generate the module-related words. Moreover, our CVLNM is flexible that we can insert other modules for specific goals. In this research, we plug a linguistic \textsc{reason} module into the decoder to approximate human-like commonsense reasoning for more descriptive captions (see Section~\ref{sec:reason_module}). 

Neural module networks have been proposed to solve various vision-language tasks, while most of them parse a module layout from a fully observable sentence, \eg, the question is used to parse a module layout in VQA~\cite{andreas2016neural}. However, for image captioning, such fully observable sentences are not available and we only have partially observable sentences during the process of captioning, which increases the difficulty of designing our CVLNM.
To address this challenge, we design a multi-head self-attention based module controller to dynamically collocate modules during \re{captioning, conditioned on the visual and linguistic context knowledge (see Section~\ref{sec:mod_col}). With this controller}, our CVLNM can discover the frequently appearing syntax patterns from the training dataset and use them to regularize the caption training and generation. During the caption generation, at each time step, we first choose a few modules and then use their outputs to generate the words. Fig.~\ref{fig:introc} sketches such captioning process where we choose the most responsible module as the example. 

Since the module layout is dynamically parsed in image captioning, it is far from perfect compared with the layout in the other visual reasoning tasks where the auxiliary sentences are available~\cite{hu2017learning,liu2019learning,shi2019explainable}. To enhance the effectiveness and robustness of our captioner, the following techniques are also applied. 1) The module controller is designed to softly fuse four distinguishable modules in the encoder to comprehensively exploit their outputs (see Section~\ref{sec:soft_fuse}). 2) A part-of-speech based syntax loss is imposed on the controller to make the generated module layout more faithful to the human-like pattern, \eg, an adjective is usually used before a noun and in such a case the syntax loss will encourage \textsc{attribute} module to be selected before \textsc{object} module (see Section~\ref{sec:ling_loss}). Furthermore, this loss encourages each module to learn its module-specific knowledge for further decomposing the image captioning task.

Extensive discussions and human evaluations are offered in Section~\ref{sec:experiment_all} to validate the effectiveness and robustness of CVLNM on the challenging MS-COCO image captioning benchmark. Overall, we achieve a new state-of-the-art 129.5 CIDEr-D score on Karpathy split and 127.8 c40 on the official server. More importantly, by following the principle of modular design, we find that our CVLNM is less likely to overfit to dataset bias. For example, compared to a strong non-module baseline Up-Down~\cite{anderson2018bottom}, our CVLNM generates 1) more accurate grammar (Fig.~\ref{fig:intro2} (a)) thanks to the joint reasoning of \textsc{function} and \textsc{object} module, 2) more descriptive attributes (Fig.~\ref{fig:intro2} (b)) due to \textsc{attribute} module, and 3) more accurate interactions (Fig.~\ref{fig:intro2} (c)) due to \textsc{relation} module. Moreover, we find that when only one training sentence of each image is provided, our CVLNM will suffer less performance deterioration compared with the strong non-module baseline. 
Our contributions can be summarized as:
\begin{itemize}[leftmargin=.1in]
\item Our CVLNM is the first module network for image captioning, which is a generic framework with principled module and controller designs. This enriches the spectrum of using neural modules for vision-language tasks.

\item We develop an advanced module controller and a syntax based loss for robustifying the dynamic module collocations and meanwhile encouraging each module to learn their module-specific knowledge. 

\item To show the extensibility of our framework, we plug a \textsc{reason} module in the decoder to approximate commonsense reasoning for more descriptive captions.

\item We conduct extensive experiments with various metrics. Experiment results show the effectiveness and robustness of our CVLNM.
\end{itemize}

This work is an extension of our previous conference paper~\cite{yang2019learning} with significant modifications:
\begin{itemize} \itemsep0pt
	\item[$\bullet$] We refined the module controller with additional inputs from previous module collocations and utilized multi-head self-attention instead of LSTM to generate the current collocation.
	\item[$\bullet$] We incorporated a novel \textsc{reason} module into the decoder to approximate commonsense reasoning for more descriptive captions.
	\item[$\bullet$] We carried detailed ablation experiments to demonstrate the effectiveness of these additional refinements. 
\end{itemize}
In addition, we re-organized the introduction, added more details, figures, explanations, results, and analyses. In particular, we analyzed the accuracy of the predicted module layout, the shape of the module distribution, and the module removal effect.

\section{Related Work}
\subsection{Image Captioning}
Most early image captioners are template-based models that first generate sentence patterns and then fill the content words like objects' categories, attributes, and their relations into the fixed patterns~\cite{kulkarni2011baby,kuznetsova2012collective,mitchell2012midge}. However, since the accuracy of the visual detectors is not satisfactory, and the template generator and the visual detectors are not jointly trained, the performances of these captioners are heavily limited. 

Compared with these template-based models, modern captioners which achieve superior performances are most attention based encoder-decoder methods~\cite{xu2015show,vinyals2015show,rennie2017self,chen2017sca,zha2019context,anderson2018bottom,yao2017boosting,luo2018discriminability,qin2019look,luo2017image}. However, unlike the template-based models, most of the encoder-decoder based models generate word one by one without explicitly exploiting syntax patterns, while various NLP works~\cite{kim2017structured,tai2015improved,shen2018ordered} have proven that syntax patterns provide beneficial inductive bias for improving robustness and effectiveness.

Our CVLNM takes \re{advantage from both} template and network based captioners, which learns hidden syntax patterns and is trained end-to-end. From the perspective of the module network, several recent captioners can be reduced to special cases of our CVLNM. For example, semantic attention based captioners~\cite{gan2017semantic,you2016image} use \textsc{object},\textsc{attribute}, and \textsc{relation} modules to predict object categories, attributes, and actions, respectively, and then directly input these semantic words into the language decoder for captioning. However, such captioners do not dynamically collocate these semantic modules and are more like extensions of the traditional attention mechanism.
KWL~\cite{lu2017knowing}, NBT~\cite{lu2018neural}, and GCN-LSTM~\cite{yao2018exploring} go one step further. They dynamically collocate \textsc{function}, \textsc{object}, and \textsc{relation} modules to generate the corresponding words, respectively. \re{Compared to} them, our CVLNM designs more fine-grained modules with diverse inductive biases and a more advanced controller for collocating modules. Some captioners also integrate different modules to generate comprehensive embeddings with various semantic knowledge. For example, Up-Down~\cite{anderson2018bottom} integrates \textsc{object} and \textsc{attribute} modules by exploiting the object and attribute labels of VG~\cite{krishna2017visual} to pre-train a Resnet-101 \re{Faster R-CNN}~\cite{ren2015faster} for extracting features; and SGAE~\cite{yang2019auto} applies Graph Neural Network to integrate \textsc{object}, \textsc{relation}, and \textsc{attribute} modules. \re{For HIP~\cite{yao2019hierarchy}, it integrates the hierarchical structure of the objects into the visual features. Thus its encoder can be considered as the integration of the object module and the relation module. }
Although such comprehensive embeddings improve the quality of the generated captions, the interpretability is also lost since we cannot figure out which module contributes more to a specific word. In contrast, our CVLNM can inspect the module fusion weights to identify which module is more responsible for a word.

\re{Some current research papers focus on developing more advanced attention techniques. For example, AoANet~\cite{huang2019attention} designs an Attention on Attention block to capture the relevance between the attention results and the queries. X-LAN~\cite{pan2020x} applies the X-Linear attention block to model the second-order interactions across multi-modal inputs. Furthermore, some researchers~\cite{li2019entangled,herdade2019image,cornia2020meshed,guo2020normalized} build their captioning models based on the Transformer architecture. For example, ETA~\cite{li2019entangled} integrates semantic and visual knowledge by the multi-head attention as the input features for captioning. Similarly, ToW~\cite{herdade2019image} integrates spatial relations and visual features for captioning. Besides exploiting the relation knowledge between objects, NSA~\cite{guo2020normalized} further normalizes self-attention. $\mathcal{M}^2$-Transformer~\cite{cornia2020meshed} designs mesh-like connections in the decoder to exploit multi-level visual features.}

\subsection{Neural Module Networks}
For visual reasoning tasks (\eg, image captioning~\cite{xu2015show} or VQA~\cite{antol2015vqa}), they are usually composed by a series of low-level vision tasks (\eg, image classification~\cite{Simonyan15}, object detection~\cite{he2016deep}, or relation recognition~\cite{lu2016visual}) and nature language processing tasks (\eg, language generation~\cite{sutskever2014sequence,vaswani2017attention} or language understanding~\cite{peters2018deep,devlin2018bert}). Then, an intuitive way to solve visual reasoning tasks is to decompose them into many simple but diverse sub-tasks and design distinguishable modules to address each sub-task.

More specifically, VQA~\cite{antol2015vqa} requires an AI agent to understand different aspects of a visual scene and the given question for correctly answering, which can be solved by a series of neural modules~\cite{andreas2016neural,andreas2016learning,hu2017learning,hu2018explainable}, \eg, \textsc{color} module for ``what is the color of an object'', \textsc{locate} module for ``where is one object'', or \textsc{count} module for ``how many is one object''. Similarly, visual reasoning~\cite{shi2019explainable,mascharka2018transparency,yi2018neural} and visual grounding~\cite{liu2018explainability,yu2018mattnet,hu2017modeling,liu2019learning} can also be decomposed into different sub-tasks addressed by specific modules. In these tasks, high-quality module layouts can be parsed from the provided fully observable sentences like the questions in VQA or the descriptions in visual grounding.

However, in image captioning, only partially observable sentences are available, and thus the module layout cannot be perfectly parsed as other vision-language tasks. To address such a challenge, we design a module controller with the previously generated module layout as input to dynamically collocate neural modules. This controller is also regularized by a part-of-speech based syntax loss for more human-like module layouts.

\section{Collocation of Visual-Linguistic Neural Modules}
\label{sec:cnm}
\begin{figure*}[t]
\centering
\includegraphics[width=1\linewidth,trim = 5mm 5mm 5mm 5mm,clip]{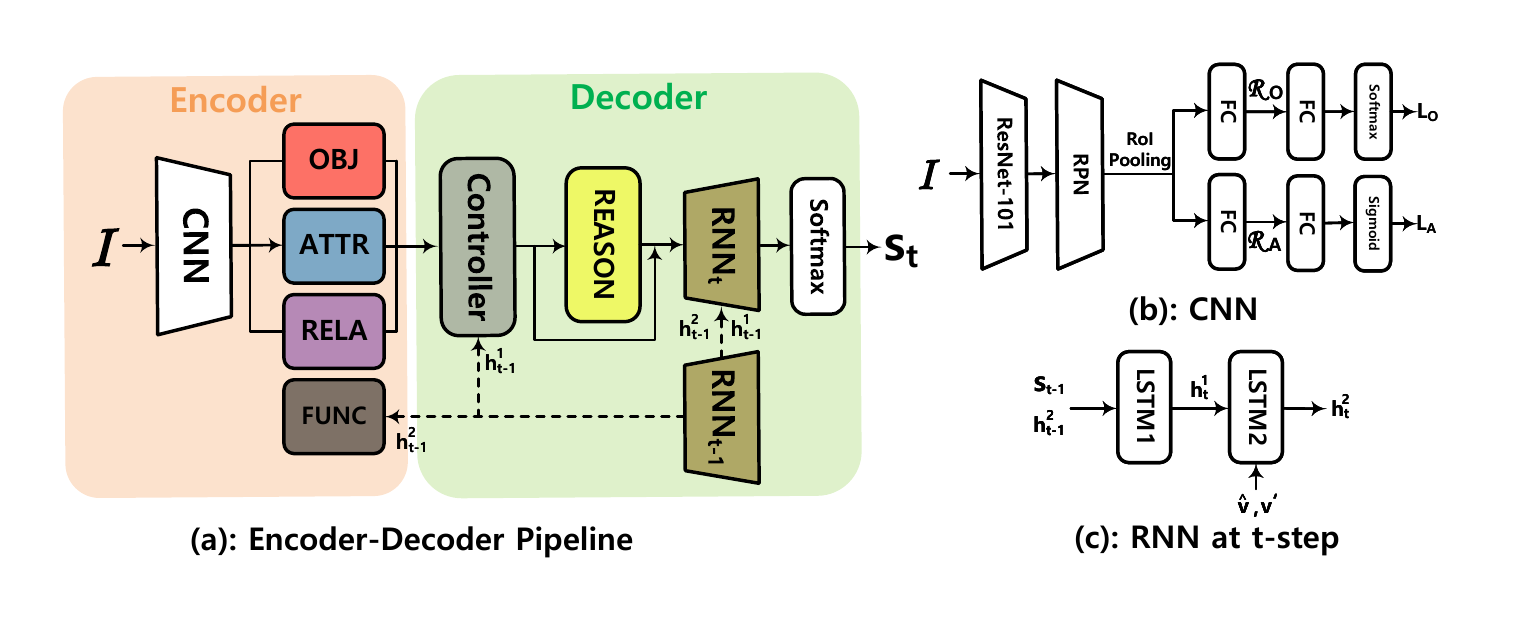}
   \caption{\re{(a) The encoder-decoder pipeline that Collocates Visual-Linguistic Neural Modules (CVLNM) for captioning an input image $\mathcal{I}$.
   The dash lines from RNN to \textsc{function} module and the module controller indicate that both sub-networks require the contextual knowledge of the partially observable sentences, which are the outputs of RNN. The self-loop in RNN denotes the recurrence structure. (b) The deployed CNN is a modified ResNet-101 Faster R-CNN. The top and bottom branches are respectively trained by object and attribute classifications. $\mathcal{R}_O$ and $\mathcal{R}_A$ are the extracted features. (c) The sketch of the Top-Down RNN (Eq.~\eqref{equ:top_down}), where $s_{t-1}$ is the last word, and $\hat{\bm{v}},\bm{v}'$ are the outputs of the module controller (Eq.~\eqref{equ:soft_fuse}) and \textsc{reason} module (Eq.~\eqref{equ:memory_network}).} 
   }
\label{fig:pip}
\end{figure*}

Fig.~\ref{fig:pip}(a) shows the structure of our learning to Collocate Visual-Linguistic Neural Modules (CVLNM) model. The encoder contains a CNN and four neural modules to extract different features (see Section~\ref{subsec:modules}). Our decoder has a module controller (see Section~\ref{sec:mod_col}) that softly fuses these features into a single feature. This feature is input into a linguistic \textsc{reason} module for commonsense reasoning (see Section~\ref{sec:reason_module}) and the output is fed into an RNN for language decoding (see Section~\ref{sec:train_infer}). We also impose a part-of-speech based syntax loss on the controller to make it more faithful to human-like syntax patterns and meantime to encourage each module to learn distinguishable knowledge (see Section~\ref{sec:ling_loss}). 

\subsection{Distinguishable Feature Extraction Modules}
\label{subsec:modules}
As sketched in the encoder of Fig.~\ref{fig:pip}(a), we use three visual and one linguistic neural modules to cover the frequently appearing part-of-speech of the captioning dataset, \ie, \textsc{object} module for nouns; \textsc{attribute} module for adjectives; \textsc{relation} module for verbs, prepositions, and quantifiers; and \textsc{function} module for other function words. 
We apply different design principles to encourage these modules to capture different aspects of information about the same image for generating different types of words, making these modules \textit{distinguishable}. For example, the output feature of \textsc{object} module should be more responsible for generating nouns instead of adjectives.

\subsubsection{\textsc{Object} Module} 
\label{subsec:object_module}
It is a visual module designed to transform the CNN features to a feature set $\mathcal{V}_O$ containing the information about object categories, \ie, $\mathcal{V}_O$ facilitates to generate nouns like ``person'' or ``dog''. The input of this module is the feature set $\mathcal{R}_O\in\mathbb{R}^{N \times d_r}$, where $N$ is the number of detected objects and $d_r$ is the dimension of an ROI feature, whose value is specified in Table~\ref{tab:parameter}. $\mathcal{R}_O$ is extracted from the top branch of Fig.~\ref{fig:pip}(b), which is a modified ResNet-101 Faster R-CNN~\cite{ren2015faster}. This branch is pre-trained by object classification using the object annotations of the VG dataset~\cite{krishna2017visual}. Specifically, it is pre-trained with the cross-entropy loss:
\begin{equation} \label{equ:obj_loss}
    L_O = -\sum_{r} \log p_{o_r},
\end{equation}
where $o_r$ is the object label of the $r$-th region and $p_{o_r}$ is the predicted probability of the object $o_r$. In this way, $\mathcal{R}_O$ will contain abundant information about the objects' categories.

Formally, this module can be written as:
\begin{equation} \label{equ:obj_mod}
\small
\begin{aligned}
 \textbf{Input:} \quad  &\mathcal{R}_O, \\
 \textbf{Output:} \quad  &\mathcal{V}_O = \text{LeakyReLU}(\text{FC}(\mathcal{R}_O)), \\
\end{aligned}
\end{equation}
where \re{FC denotes the fully-connected layer, LeakyReLU denotes the Leak ReLU layer~\cite{xu2015empirical},} $\mathcal{V}_O\in\mathbb{R}^{N \times d_v}$ is the output feature set, and $d_v$ is the feature dimension whose value is specified in Table~\ref{tab:parameter}.

\subsubsection{\textsc{Attribute} Module} 
\label{subsec:attribute_module}
It is a visual module designed to transform the CNN features to a feature set $\mathcal{V}_A$ containing attribute information for generating adjectives like ``black'' or ``small''. The input of this module is $\mathcal{R}_A \in \mathbb{R}^{N \times d_r}$,
which is extracted from the bottom branch of Fig.~\ref{fig:pip}(b). This branch is pre-trained by multi-label attribute classification using the attribute annotations of VG with the binary cross-entropy loss:
\begin{equation} \label{equ:attr_loss}
    L_A = -\sum_{r} \sum_{i} a_{r,i} \log p_{a_{r,i}},
\end{equation}
where $a_{r,i}$ is a binary variable denoting whether the $r$-region has the $i$-th attribute and $p_{a_{r,i}}$ is the predicted possibility of the $i$-th attribute. 
After pre-training, $\mathcal{R}_A$ will contain abundant information about objects' attributes. Formally, this module can be written as:
\begin{equation} \label{equ:attr_mod}
\small
\begin{aligned}
 \textbf{Input:} \quad  &\mathcal{R}_A, \\
 \textbf{Output:} \quad  &\mathcal{V}_A=\text{LeakyReLU}(\text{FC}(\mathcal{R}_A)), \\
\end{aligned}
\end{equation}
where $\mathcal{V}_A \in \mathbb{R}^{N \times d_v}$ is the output of this module.

\subsubsection{\textsc{Relation} Module} 
\label{subsec:relation_module}
It is a visual module designed to transform CNN features to a feature set $\mathcal{V}_R$ containing information about potential relations between objects, which would help generate verbs like ``ride'', prepositions like ``on'', or quantifiers like ``three''. This module is built upon the multi-head self-attention mechanism~\cite{vaswani2017attention}, which automatically seeks the relations among the objects during the attention computations between the corresponding features. Here, we use $\mathcal{R}_O$ in Eq.~\eqref{equ:obj_mod} as the input because these kinds of features are widely applied as the inputs for successful relation detections~\cite{zhang2017visual,zellers2018neural}. This module is formulated as:

\begin{equation} \label{equ:rela_mod}
\small
\begin{aligned}
 \textbf{Input:} \quad  &\mathcal{R}_O, \\
 \textbf{Attention:} \quad & \mathcal{M}=\text{MH-ATT}(\mathcal{R}_O), \\
 \textbf{Output:} \quad  &\mathcal{V}_R=\text{LeakyReLU}(\text{MLP}(\mathcal{M})), \\
\end{aligned}
\end{equation}
where MH-ATT($\cdot$) is the multi-head attention mechanism, when the input query, key, and value are the same, it is also called multi-head self-attention; MLP($\cdot$) is a feed-forward network containing two fully connected layers with a ReLU activation layer in between~\cite{vaswani2017attention}: FC-ReLU-FC; and $\mathcal{V}_R \in \mathbb{R}^{N \times d_v}$ is the output of this module. Specifically, the following steps are deployed to compute the MH-ATT. We first use the scaled dot-product to compute $k$ head matrices \textbf{head}$_i\in \mathbb{R}^{N \times d_k}$:
\begin{equation}
\label{equ:rela_self_attention}
\small
    \textbf{head}_i=\text{Softmax}( \frac{\mathcal{R}_O\bm{W}_i^1(\mathcal{R}_O\bm{W}_i^2)^T}{\sqrt{d_k}} )\mathcal{R}_O\bm{W}_i^3,
\end{equation} 
where $\bm{W}_i^1, \bm{W}_i^2, \bm{W}_i^3 \in \mathbb{R}^{d_r \times d_k}$ are all trainable matrices, $d_k=d_r/k$ and $k$ is the number of head matrices. Then these $k$ head matrices are concatenated and linearly projected to the feature set $\mathcal{M}\in \mathbb{R}^{N \times d_r}$:
\begin{equation}
\label{equ:multi_head}
\small
    \mathcal{M} = [\textbf{head}_1,...,\textbf{head}_k]\bm{W}_{C},
\end{equation}
where $[\cdot]$ means the concatenation operation and $\bm{W}_{C} \in \mathbb{R}^{d_r\times d_r}$ is a trainable matrix. 

\begin{figure}[t]
\centering
\includegraphics[width=1\linewidth,trim = 5mm 5mm 5mm 5mm,clip]{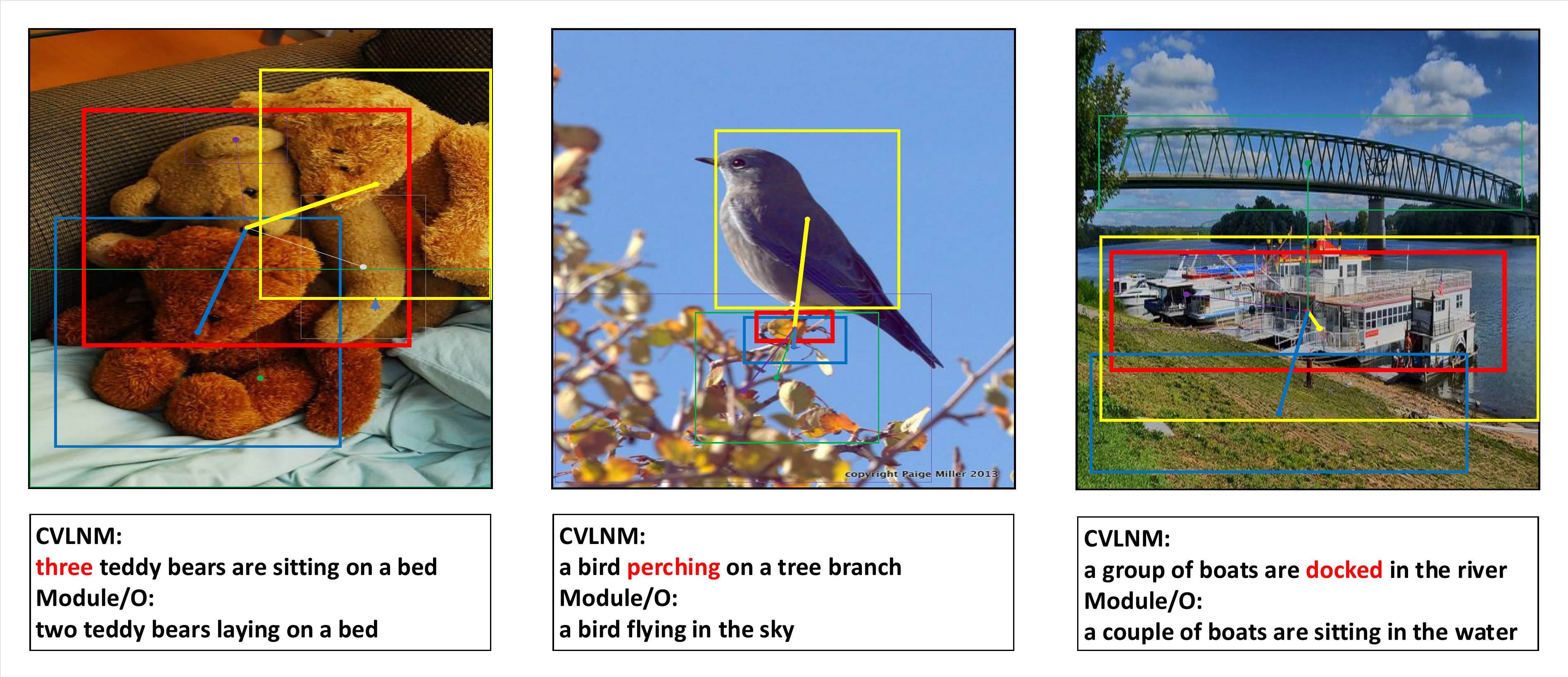}
  \caption{Three examples show how \textsc{relation} module generates relation specific words. The red box is the attended region when \textsc{relation} module generates a relation specific word. The thickness of the line connecting two boxes is determined by the soft attention weight computed by Eq.~\eqref{equ:rela_self_attention}, the larger the attention weight, the thicker the line.}
\label{fig:supp_rela}
\end{figure}
Fig.~\ref{fig:supp_rela} demonstrates how this self-attention based \textsc{relation} module generates relation specific words. For example, in the middle figure, at the third time step, \textsc{relation} module focuses more on the ``paw'' part (red box) of the bird, and meantime the clues about ``bird'' (yellow box) and ``tree'' (blue box) are incorporated to the feature of the ``paw'' part by MH-ATT. By exhaustively considering these visual clues, a more accurate action ``perch'' is generated by our \textsc{relation} module.

\subsubsection{\textsc{Function} Module} 
It is a linguistic module designed to produce a single feature $\hat{\bm{v}}_F$ for generating function words like ``a'' or ``and''. The input of this module is the linguistic context vector $\bm{h}_t^2 \in \mathbb{R}^{d_h}$ accumulated in the RNN, which is the output of the second LSTM in Fig.~\ref{fig:pip}(c).
$d_h$ is the dimension of this context vector, whose value is given in Table~\ref{tab:parameter}. We use $\bm{h}_t^2$ as the input because it contains rich linguistic context knowledge of the partially generated captions and such knowledge instead of visual knowledge is more beneficial for generating function words. This module is formulated as:
\begin{equation} \label{equ:func_mod}
\small
\begin{aligned}
 \textbf{Input:} \quad  &\bm{h}_t^2, \\
 \textbf{Output:} \quad  &\hat{\bm{v}}_F=\text{LeakyReLU}(\text{FC}(\bm{h}_t^2)), \\
\end{aligned}
\end{equation}
where $\hat{\bm{v}}_F \in \mathbb{R}^{d_v}$ is the output feature.

\subsection{Module Controller}
\label{sec:mod_col}
\begin{figure}[t]
\centering
\includegraphics[width=1\linewidth,trim = 5mm 5mm 5mm 5mm,clip]{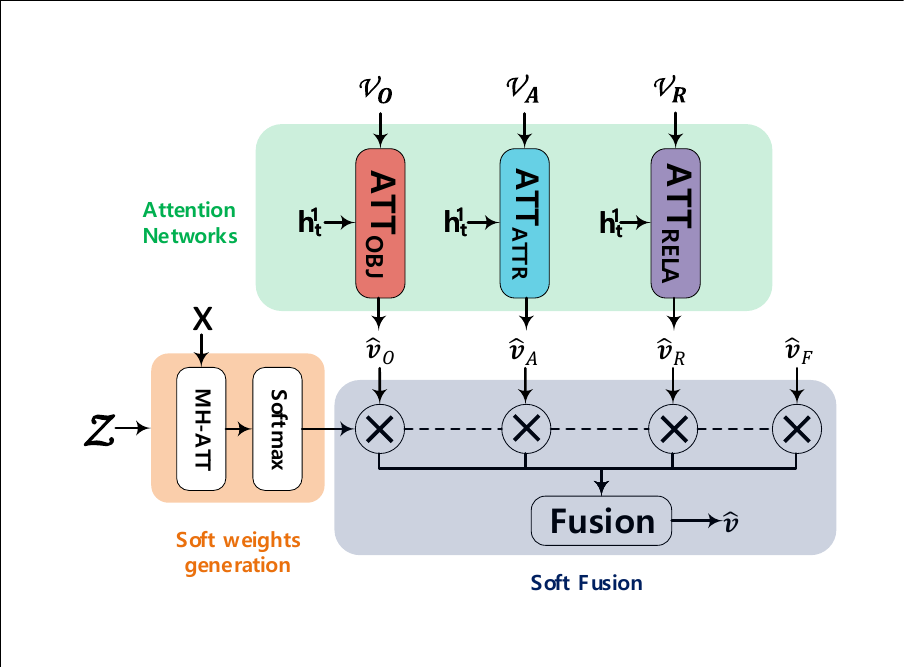}
   \caption{
   Our module controller has three matrix-sum attention (MS-ATT) networks (ATT$_\textsc{obj}$, ATT$_\textsc{attr}$, ATT$_\textsc{rela}$) and a multi-head attention network (MH-ATT). MH-ATT generates four soft weights to fuse the four attended features ($\hat{\bm{v}}_O$, $\hat{\bm{v}}_A$, $\hat{\bm{v}}_R$, $\hat{\bm{v}}_F$) into a single feature $\hat{\bm{v}}$. $\bm{h}_t^1$ (Eq.~\eqref{equ:att_nets}) is the query vector used in three MS-ATT networks, $\mathcal{Z}$ (Eq.~\eqref{equ:z_mhsa}) is the embedding set of the previous module collocations, and $\bm{x}$ (Eq.~\eqref{equ:equ_x}) is the query vector used in MH-ATT.
   }
\label{fig:cont}
\end{figure}
\re{It is still an open question on how to define a perfect complete set of neural modules for visual reasoning~\cite{yu2018mattnet,andreas2016neural}. We} believe that a variety of complex tasks can be approximately accomplished by adaptively selecting modules from an empirically complete module set conditioned on the visual and linguistic context~\cite{hu2018explainable}. Motivated by this, we design a module controller for dynamically collocating the modules when only a partially generated caption is available. Fig.~\ref{fig:cont} shows the detail of this controller, which contains three matrix-sum attention (MS-ATT) networks and one multi-head attention (MH-ATT) network based soft-weight generator. The fused feature vector $\hat{\bm{v}}$ is the output of this controller, which will be input into the followed \textsc{reason} module and RNN for the next step reasoning (Fig.~\ref{fig:pip}). Next, we describe how each component of our module controller works.

\subsubsection{Module Attention}
Before the soft fusion, three MS-ATT are used to respectively transform three visual modules' output into three more informative features:
\begin{equation} \label{equ:att_nets}
\small
\begin{aligned}
 &\textbf{ATT}_{OBJ}:  \quad  &\hat{\bm{v}}_O = \text{MS-ATT}(\mathcal{V}_O,\bm{h}_t^1), \\
 &\textbf{ATT}_{ATTR}: \quad  &\hat{\bm{v}}_A = \text{MS-ATT}(\mathcal{V}_A,\bm{h}_t^1), \\
 &\textbf{ATT}_{RELA}: \quad  &\hat{\bm{v}}_R = \text{MS-ATT}(\mathcal{V}_R,\bm{h}_t^1), \\
\end{aligned}
\end{equation}
where $\hat{\bm{v}}_O,\hat{\bm{v}}_A,\hat{\bm{v}}_R\in \mathbb{R}^{d_v}$ are the transformed features of $\mathcal{V}_O$, $\mathcal{V}_A$, $\mathcal{V}_R$, which are outputs of \textsc{object}, \textsc{attribute}, \textsc{relation} modules (Section~\ref{subsec:modules}), respectively; $\bm{h}_t^1\in \mathbb{R}^{d_h}$ is the query vector produced by the first LSTM in the RNN as sketched in Fig.~\ref{fig:pip}(c); and three MS-ATT networks own the same structure while the parameters are not shared:
\begin{equation} \label{equ:attend}
\small
\begin{aligned}
 \textbf{Input:} \quad  &\mathcal{V}, \bm{h},\\
 \textbf{Attention:}  \quad  &
 \begin{aligned} &a_i=\bm{\omega}_a^T\tanh({\bm{W}_v\bm{v}_i+\bm{W}_h\bm{h}}),\\
  &\bm{\alpha}=\text{softmax}(\bm{a}), 
  \end{aligned}
  \\
 \textbf{Output:} \quad  &\hat{\bm{v}}=\mathcal{V}\bm{\alpha}, \\
\end{aligned}
\end{equation}
where $\bm{W}_h\in\mathbb{R}^{d_a\times d_h}$, $\bm{W}_v\in\mathbb{R}^{d_a\times d_v}$ and $\bm{\omega}_a\in\mathbb{R}^{d_a}$ are all trainable parameters. 

\subsubsection{Soft Fusion}
\label{sec:soft_fuse}
Since we only have the partially observable caption to parse the module layout, the parsed layout is far from perfect. To enhance the robustness\footnote{\re{``Robustness'' means that the soft fusion strategy helps generate more ``accurate'' captions.}}, we softly fuse these module outputs to provide more comprehensive vision and language knowledge. After computing the features from each module: $\hat{\bm{v}}_O$, $\hat{\bm{v}}_A$, $\hat{\bm{v}}_R$ (Eq.~\eqref{equ:att_nets}),and $\hat{\bm{v}}_F$ (Eq.~\eqref{equ:func_mod}), the controller generates four soft weights to fuse them. 

Since the module collocation at each time step is highly related to the previous module collocations, we build this generator upon the MH-ATT network with the previous module collocations $\mathcal{Z}$ as input. We apply MH-ATT instead of LSTM (which is used in our preliminary work~\cite{yang2019learning}) since the current module collocation is possibly related to a more remote module collocation and MH-ATT is good at dealing with such long-range dependencies~\cite{vaswani2017attention,kitaev2018constituency}. 

At time step $t$, $\mathcal{Z}=\{\bm{z}_{0:t-1}\}\in\mathbb{R}^{t \times d_z}$ where $\bm{z}_0$ is the embedding of the start token and $\bm{z}_t$ is the embedding of the chosen modules at time step $t$ (Eq.~\eqref{equ:equ_module_embedding}). We first deploy MH-ATT upon $\mathcal{Z}$ to build the connections among collocations in different steps:
\begin{equation} \label{equ:z_mhsa}
    \hat{\mathcal{Z}} = \text{MH-ATT}(\mathcal{Z}),
\end{equation}
where MH-ATT($\cdot$) has the similar formula as Eq.~\eqref{equ:rela_self_attention} and~\eqref{equ:multi_head} with different parameters. Then the multi-modal context vector $\bm{x}\in\mathbb{R}^{d_z}$ is used as the query vector to get the current soft collocation weights and $\bm{x}$ is:
\begin{equation}
\label{equ:equ_x}
    \bm{x}=\text{ReLU}(\text{FC}([\hat{\bm{v}}_O,\hat{\bm{v}}_A,\hat{\bm{v}}_R,\bm{h}_t^2])),
\end{equation}
where $\bm{h}_t^2$ is the context vector output by the second LSTM of the RNN as shown in Fig.~\ref{fig:pip}(c). We use $\bm{x}$ as the query vector because both the visual clues ($\hat{\bm{v}}_O,\hat{\bm{v}}_A,\hat{\bm{v}}_R$) and the linguistic context ($\bm{h}_t^2$) of the partially generated caption are indispensable for successfully collocating modules. The process of soft fusion is formulated as:
\begin{equation} \label{equ:soft_fuse}
\small
\begin{aligned}
 \textbf{Input:} \quad  &\hat{\mathcal{Z}},\bm{x}, \\
 \textbf{MH-ATT:} \quad  &
 \begin{aligned} &\textbf{head}_i=\text{Softmax}( \frac{\bm{x}\bm{W}_l^1(\hat{\mathcal{Z}}\bm{W}_l^2)^T}{\sqrt{d_z}} )\hat{\mathcal{Z}}\bm{W}_l^3, \\
  &\hat{\bm{x}}=[\textbf{head}_1,...,\textbf{head}_j]\bm{W}_Z,
  \end{aligned}
  \\
 \textbf{Soft Vector:} \quad  &\bm{w} = \{w_O,w_A,w_R,w_F\}=\text{Softmax}(\text{FC}(\hat{\bm{x}})), \\
 \textbf{Output:} \quad  &\hat{\bm{v}}=[w_O \hat{\bm{v}}_O,w_A \hat{\bm{v}}_A,w_R \hat{\bm{v}}_R,w_F \hat{\bm{v}}_F], \\
\end{aligned}
\end{equation}
where $\bm{W}_l^1, \bm{W}_l^2, \bm{W}_l^3\in\mathbb{R}^{d_z \times d_j}$, $\bm{W}_{Z}\in\mathbb{R}^{d_z \times d_z}$ are all trainable matrices, $d_j=d_z/j$ is the dimension of each head vector, and $j$ is the number of the head vectors; \re{$w_O$,$w_A$, $w_R$,$w_F$ are scalar fusion weights}; and the output fusion vector is $\hat{\bm{v}}\in\mathbb{R}^{4d_v}$.

After generating the fusion weights $\bm{w}$, we compute the $t$-th module embedding $\bm{z}_t$ as:
\begin{equation}
\label{equ:equ_module_embedding}
    \bm{z}_t=w_O\bm{e_O}+w_A\bm{e_A}+w_R\bm{e_R}+w_F\bm{e_F},
\end{equation}
where $\bm{e_O},\bm{e_A},\bm{e_R},\bm{e_F}\in\mathbb{R}^{d_z}$ are four learnable label embeddings\footnote{Learnable label embeddings mean that they are 4-dimensional one hot vectors multiplied with a learnable $4\times d_z$ matrix.} corresponding to \textsc{object}, \textsc{attribute}, \textsc{relation}, and \textsc{function} modules, respectively. The positional encoding~\cite{vaswani2017attention} is also used to incorporate the position knowledge of the module layout, whose size is set to $d_z$. This position encoding and $\bm{z}_t$ are summed together to get the module embedding, which is used in Eq.~\eqref{equ:z_mhsa}, while for convenience, we still use $\bm{z}_t$ to denote it. 

\subsection{Syntax Loss}
\label{sec:ling_loss}
To further encourage each module to learn distinguishable knowledge, \eg, \textsc{object} module focuses more on object categories instead of visual attributes, and make the module layout faithful to human-like collocations, \eg, adjectives are usually be said before nouns, we design a part-of-speech based syntax loss and impose it on the module controller.

We build this loss by extracting the words' part-of-speech (\eg, adjectives, nouns, or verbs) from the ground-truth captions by a part-of-speech tagger tool~\cite{toutanova2000enriching}. According to the extracted part-of-speech, we assign each word an one hot vector
$\bm{w}^{*}=\{w_O^{*},w_A^{*},w_R^{*},w_F^{*}\}$, indicating which module should be chosen for generating this word. In particular, we assign \textsc{object} module to nouns (NN like ``bus'') by setting $w_O^{*}=1$, \textsc{attribute} module to adjectives (ADJ like ``green'') by setting $w_A^{*}=1$; \textsc{relation} module to verbs (VB like ``drive''), prepositions (PREP like ``on''), and quantifiers (CD like ``three'') by setting $w_R^{*}=1$; and \textsc{function} module to the other words (CC like ``and'') by setting $w_F^{*}=1$.

Given these expert-guided module tags $\bm{w}^{*}$, our syntax loss $L_{s}$ is defined as the cross-entropy value between $\bm{w}^{*}$ and soft fusion weights $\bm{w}$ calculated from Eq.~\eqref{equ:soft_fuse}:
\re{
\begin{equation}
\small
\label{equ:ling_loss}
    L_{s}=-(w_O^{*} \log w_O + w_A^{*} \log w_A + w_R^{*} \log w_R+ w_F^{*} \log w_F).
\end{equation}
}
Importantly, when this loss is imposed on the controller, the soft fusion strategy in Section~\ref{sec:soft_fuse} approximates to the hard selection that only one module will be responsible for the generated word. For example, when the noun ``dog'' is the supervision, only $w_{O}^*=1$ and $w_{O}$ in Eq.~\eqref{equ:soft_fuse} is encouraged to be 1 due to this syntax loss. Then the $\hat{\bm{v}}_O$ part in the fusion representation $\hat{\bm{v}}$ will get more back-propagated gradients from the noun ``dog'', and thus \textsc{object} module is more encouraged to learn from this noun. In this way, this syntax loss regularizes the training of the modules to make them more distinguishable. Such argument is validated in Section~\ref{sec:experiment} where we find that this loss indeed increases the recall of each module-specific word and the accuracy of the module collocation.

\subsection{\textsc{Reason} Module}
\label{sec:reason_module}
In addition to distinguishing different aspects of an image during captioning, humans can also achieve commonsense reasoning to summarize or even infer something that is not evidently detected from the image. For example, when we observe some persons holding umbrellas on a wet road, we can infer that it is raining even the raindrops are not obvious (which means \textsc{object} module can hardly extract evident information about the object ``raindrop''). To equip our captioner with such commonsense reasoning, we plug a linguistic \textsc{reason} module into the decoder as sketched in Fig.~\ref{fig:pip}(a). This module is inserted between the controller and the RNN to imitate that humans usually achieve commonsense reasoning after grasping the comprehensive knowledge of a visual scene, while such knowledge is preserved in the fused representation $\hat{\bm{v}}$ (Eq.~\eqref{equ:soft_fuse}). 

We exploit the ConceptNet~\cite{liu2004conceptnet} dataset, which is built for helping computers to achieve commonsense reasoning, to build our \textsc{reason} module. We collect the relation triplets from this dataset and embed them into a key-value memory network~\cite{sukhbaatar2015end,miller2016key}: $\mathcal{M}=\{\bm{m}_1, \bm{m}_2,..., \bm{m}_K\}\in \mathbb{R}^{d_m \times K}$, where each $\bm{m}_k$ is the embedding of a relation triplet, which is represented as ``subject-predicate-ob\-ject'', \eg, ``eagle-type of-bird'' or ``umbrella-related to-rain''.We use an efficient relation embedding operation to get $\bm{m}$:
\begin{equation}\label{equ:equ_relation_embedding}
    \bm{m}=\text{ReLU}(\text{FC}([\bm{e}_s, \bm{e}_p, \bm{e}_o])),
\end{equation}
where $\bm{e}_s, \bm{e}_p, \bm{e}_o \in \mathbb{R}^{d_e}$ denote trainable word embeddings of the subject, predicate, and object, respectively. 

Given this memory network, we apply dot-product attention (DP-ATT) to imitate commonsense reasoning: the fused representation $\hat{\bm{v}}$ acts as the query and the embedding $\bm{m}_k$ acts as both the key and the value:
\begin{equation} \label{equ:memory_network}
\small
\begin{aligned}
 \textbf{Input:} \quad  &\hat{\bm{v}}, \mathcal{M}=\{\bm{m}_1, \bm{m}_2,..., \bm{m}_K\} \\
 \textbf{Attention:} \quad  &\bm{\beta} = \text{softmax}(\mathcal{M}^T\bm{W}_v\hat{\bm{v}}), \\
 \textbf{Output:} \quad  &\bm{v}'=\mathcal{M} \bm{\beta}=\sum_{k=1}^K\beta_k\bm{m}_k, \\
\end{aligned}
\end{equation}
where $\bm{W}_v \in \mathbb{R}^{d_m \times 4d_v}$ is a trainable matrix and $\bm{v}'$ is the output of \textsc{reason} module, which provides commonsense knowledge for the RNN to captioning. 

\subsection{Training and Inference of CVLNM}
\label{sec:train_infer}
As shown in Fig.~\ref{fig:pip}, by using Faster R-CNN~\cite{he2016deep}, five neural modules, the module controller, and Top-Down RNN~\cite{anderson2018bottom}, our CVLNM can predict the probability of the next word $s_t$ given the image $\mathcal{I}$ and the last word $s_{t-1}$. Specifically, as shown in Fig.~\ref{fig:pip}(c), Top-Down RNN contains two LSTM layers and it generates the word distribution $P(s_t)$ as:
\begin{equation} \label{equ:top_down}
\small
\begin{aligned}
 \textbf{Input:} \quad  &\bm{s}_{t-1}, \hat{\bm{v}},\bm{v}' \\
 \textbf{LSTM1:}  \quad  & \bm{h}_{t}^1=\text{LSTM1}([\text{Embed}(\bm{s}_{t-1}),\bm{h}_{t-1}^2]) \\
 \textbf{LSTM2:}  \quad  & \bm{h}_{t}^2=\text{LSTM2}([\hat{\bm{v}},\bm{v}',\bm{h}_{t}^1]) \\
 \textbf{Output:} \quad  & P(s_t)=\text{Softmax}(\text{FC}(\bm{h}_{t}^2)), \\
\end{aligned}
\end{equation}
where $\hat{\bm{v}},\bm{v}'$ are the outputs of the module controller (Eq.~\eqref{equ:soft_fuse}) and \textsc{reason} module (Eq.~\eqref{equ:memory_network}); Embed$(\cdot)$ is the learnable word embedding layer; LSTM1$(\cdot)$ and LSTM2$(\cdot)$ are two different LSTM layers.

Given the ground-truth caption $\mathcal{S}^{*}=\{s_{1:T}^{*}\}$ with part-of-speech tags $\mathcal{W}^{*}=\{\bm{w}_{1:T}^{*}\}$, we can end-to-end train our CVLNM by minimizing the syntax loss $L_{s}$ defined in Eq.~\eqref{equ:ling_loss} and the language loss $L_{l}$. Given the predicted word distribution $P(s)$, we can define the language loss $L_{l}$ as the cross-entropy loss:
\begin{equation}
\small
     L_{l}=L_{XE} = -\sum_{t=1}^T \log P(s_t^*),
\label{equ:equ_celoss}
\end{equation}
or the negative reinforcement learning (RL) based reward~\cite{rennie2017self}:
\begin{equation}
\small
    L_{l}=L_{RL} = -\mathbb{E}_{s_{t}^s \sim P(s)}[r(s_{1:T}^s;s_{1:T}^*)],
    \label{equ:equ_rlloss}
\end{equation}
where $r$ is a sentence-level metric, \eg, CIDEr-D~\cite{vedantam2015cider}, between the sampled sentence $\mathcal{S}^s=\{s_{1:T}^s\}$ and the ground-truth $\mathcal{S}^{*}=\{s_{1:T}^{*}\}$. Then the total loss becomes
\begin{equation}
\small
    L = L_{l} + \lambda L_{s},
\label{equ:total_loss}
\end{equation}
where $\lambda$ is the \re{weighting of two objectives}. During inference, we adopt the beam search strategy~\cite{rennie2017self} with a beam size of 5 to sample the caption from the predicted distribution $P(s)$.

\section{Experiments}
\label{sec:experiment_all}
\subsection{Datasets and Settings}
\label{subsec:dataset}
\noindent\textbf{MS-COCO~\cite{lin2014microsoft}} has an official split: 82,783/40,504/40,775 images for training/validating/testing, respectively. The 3rd-party Karpathy split~\cite{karpathy2015deep} provides an off-line test, which has 113,287/5,000/5,000 images for training/validating/testing, respectively. We deployed our models on both splits to validate the effectiveness. The captions were pre-processed by the following steps: the texts were tokenized on white spaces; all the letters were changed to lowercase; the words were removed if they appear less than 5 times and then we had a vocabulary with $10,369$ words; each caption was trimmed to a maximum of $16$ words.

\noindent\textbf{Visual Genome~\cite{krishna2017visual} (VG)} 
is a noisy scene graph dataset that a large number of the object and attribute labels only appear in a few fractions of the whole training annotations. Thus it is a common practice to filter the dataset for better usage. For example, in most scene graph detection models like~\cite{xu2017scene,yang2018graph,zellers2018neural}, researchers only use 150 objects and 50 relations to train their scene graph detectors. Researchers in the field of image captioning also filter this dataset, \eg, Up-Down~\cite{anderson2018bottom} uses a subset with 1600 objects and 400 attributes. Following them, we also filtered this dataset by keeping the labels that appear more than $2,000$ times in the training set, which results in 305 objects and 103 attributes remaining. Importantly, since some images co-exist in both VG and COCO, we filtered out the images and their annotations of VG which appear in the COCO test set. When pre-training the CNN shown in Fig.~\ref{fig:pip}(b), we sequentially minimized $L_O$ (Eq.~\eqref{equ:obj_loss}) and $L_A$ (Eq.~\eqref{equ:attr_loss}) using the object and attribute annotations of VG. After pre-training, we extracted the features as the inputs into different modules.

\noindent\textbf{ConceptNet~\cite{liu2004conceptnet}} is a freely-available semantic dataset containing abundant commonsense triplets formed as ``subject-predicate-object''. Each triplet is assigned with an important weight. We exploited these weighted triplets to build our \textsc{reason} module (Section~\ref{sec:reason_module}) for approximating commonsense reasoning. We searched the related triplets from ConceptNet using the words in caption vocabulary as the keys and preserved the top 10,000 searched commonsense triplets into the memory network $\mathcal{M}$ according to the importance weight of each triplet. 

\noindent\textbf{Settings.}
\begin{table}[t]
\begin{center}
\caption{The number of trainable parameters.}
\label{tab:parameter}
\begin{tabular}{c c c | c c c}
		\hline
		   symbol   & equation & number & symbol   & equation & number\\ \hline
           $d_r$ & Eq.~\eqref{equ:obj_mod} & 2048 & $d_v$ & Eq.~\eqref{equ:obj_mod} & 1000\\
           $k$ & Eq.~\eqref{equ:rela_self_attention} & 8 & $d_k$ & Eq.~\eqref{equ:rela_self_attention} & 256\\
           $d_h$ & Eq.~\eqref{equ:func_mod} & 1000 & $d_a$ & Eq.~\eqref{equ:attend} & 512\\
           $d_z$ & Eq.~\eqref{equ:soft_fuse} & 1000 & $j$ & Eq.~\eqref{equ:soft_fuse} & 8\\
           $d_j$ & Eq.~\eqref{equ:soft_fuse} & 125 &$d_e$ & Eq.~\eqref{equ:equ_relation_embedding} & 1000\\
           $d_m$ & Eq.~\eqref{equ:memory_network} & 1000 &  $K$ & Eq.~\eqref{equ:memory_network} & 10,000\\
           \hline
\end{tabular}
\end{center}
\end{table}
Table~\ref{tab:parameter} summarizes \re{the number of training parameters in each module}. We used Adam optimizer~\cite{kingma2014adam} to train the whole model. The learning rate was initialized to $5e^{-4}$ and was decayed by $0.8$ for every $5$ epochs. $L_{XE}$ Eq.~\eqref{equ:equ_celoss} and $L_{RL}$ Eq.~\eqref{equ:equ_rlloss} were in turn used as the language loss to train our CVLNM for 35 epochs and 65 epochs, respectively. In the experiments, we found that the performance is non-sensitive to $\lambda$ in Eq.~\eqref{equ:total_loss}. By default, we empirically set $\lambda=1$ and $\lambda=0.5$ when $L_{XE}$ and $L_{RL}$ were used as the language loss, respectively. The batch size was set to 100. 

\subsection{Ablation Studies}
\label{sec:experiment}
We conducted extensive ablations to confirm the effectiveness of each component by gradually incorporating them into the pipeline, including four distinguishable modules in the encoder (see Section~\ref{subsec:modules}), the multi-head attention (MH-ATT) based module controller (see Section~\ref{sec:mod_col}), the part-of-speech based syntax loss (see Section~\ref{sec:ling_loss}), and \textsc{reason} module (see Section~\ref{sec:reason_module}). We arrange the ablation studies by proposing research questions (\textbf{Q}) in Section~\ref{subsec:research_questions}, specifying miscellaneous metrics in Section~\ref{sec:metrics}, and providing the corresponding empirical answers (\textbf{A}) in Section~\ref{subsec:empirical_answers}.

\subsubsection{Research Questions and the Corresponding Baselines}
\label{subsec:research_questions}
\noindent\textbf{Q1:} Will each visual feature extraction module introduced in Section~\ref{subsec:modules}, \ie, \textsc{object}, \textsc{attribute}, and \textsc{relation} modules, learn distinguishable knowledge and then generate more accurate module-specific words, \eg, will \textsc{object} module generate more accurate nouns? Does CVLNM allow controllable caption generation, \eg, will removing \textsc{attribute} module eliminate all the attribute related words?

To answer them, the following baselines were designed: we only used a single visual module in the encoder to extract features and deployed the Top-Down RNN~\cite{anderson2018bottom} as the decoder. When \textsc{object}, \textsc{attribute}, and \textsc{relation} modules were used, the baselines are denoted as \textbf{Module/O}, \textbf{Module/A}, and \textbf{Module/R}, respectively. Note that the baseline Module/O is an upgraded version of the Up-Down model~\cite{anderson2018bottom}. After training the model with all modules, during testing, we cut off a module and then measure the ratio of the corresponding words to all the generated words to inspect the removal effect.

\noindent\textbf{Q2:} Will the qualities of the generated captions be improved when the modules in the encoder are fused? How to fuse them to achieve a better performance?

To answer them, the following three strategies were designed, where each one used a specific type of fusion weights. \textbf{Col/1} denotes the model which sets all the fusion weights as 1. This is equivalent to using all the module features in the Up-Down model. When learnable soft fusion weights $\bm{w}$ (Eq.~\eqref{equ:soft_fuse}) were used, the method is called \textbf{Col/S}. When the Gumbel-Softmax layer~\cite{jang2016categorical} was used to transfer $\bm{w}$ into one hot vector for achieving hard selection, the baseline is called \textbf{Col/H}. In both Col/S and Col/H, we input the previous module layout $\mathcal{Z}$ into the MH-ATT based controller to get $\bm{w}$.

\noindent\textbf{Q3:} Will inputting the previous module layout $\mathcal{Z}$ (Eq.~\eqref{equ:soft_fuse}) into the controller generate better soft fusion weights? Will MH-ATT preserve more long-range dependencies between remote module collocations than LSTM (which is deployed in our preliminary work CNM~\cite{yang2019learning})? More importantly, will better soft fusion weights encourage each module to learn its distinguishable knowledge?

To answer them, we compared the following baselines. When we built the controller upon MH-ATT with/without $\mathcal{Z}$, the baselines are named as \\textbf{MH-ATT}+$\mathcal{Z}$/\textbf{MH-ATT}. In the same vein, when the module controller is built upon LSTM, the baselines are \textbf{LSTM+$\mathcal{Z}$}/\textbf{LSTM}. In these baselines, the encoder contains all the four modules, the soft fusion strategy was applied, and we did not use \textsc{reason} module and the syntax loss. Note that MH-ATT+$\mathcal{Z}$ equals the baseline Col/S in Q2. 

\noindent\textbf{Q4:} Will the expert part-of-speech knowledge provided by the syntax loss $L_{s}$ (Eq.~\eqref{equ:ling_loss}) benefit the model? Will such loss encourage each feature extraction module to learn more distinguishable knowledge?

To answer them, we compared the models by training them with and without $L_{s}$. When we used $L_{s}$ in MH-ATT+$\mathcal{Z}$, we have \textbf{MH-ATT}+$\mathcal{Z}$+$L_{s}$. 

\noindent\textbf{Q5:} Will \textsc{reason} module generate more human-like captions? 

To answer it, we inserted \textsc{reason} module into MH-ATT+$\mathcal{Z}$ to get \textbf{MH-ATT}+$\mathcal{Z}$+Reason and compared them.

\noindent\textbf{Q6:} Will the integrated model achieve the best performances among all the baselines?  

We compared our full \textbf{CVLNM} with the other baselines. 

\noindent\textbf{Q7:} What do the soft weights look like in each step, \eg, are they sharp (putting most weights on one module) or flat? Will each word be generated mostly from a single module or multiple modules?

To answer this, for each word, we computed the average entropy of the module distribution and the average probability of the most responsible module, where ``the most responsible'' denotes that this module has the largest weight when one word is generated.

\noindent\textbf{Q8:} How about changing the hyperparameters, \eg, $\lambda$ in Eq.~\eqref{equ:total_loss} and the number of annotations for training our CNN? 

We set $\lambda$ to different fixed values when $L_{XE}$ and $L_{RL}$ were used to train the model, which is named \textbf{CVLNM ($\lambda$)}. We used 1600/400 objects/attributes as Up-Down~\cite{anderson2018bottom} to pre-train the feature extractor CNN. The model using these features is denoted as \textbf{CVLNM+}.

\subsubsection{Evaluation Metrics for Answering Questions}
\label{sec:metrics}
To quantitatively answer the above research questions, we applied the following metrics to comprehensively test the effectiveness and robustness of our CVLNM. 

1) We measured the similarities between the generated captions and the ground-truth captions by five standard metrics, which are CIDEr-D~\cite{vedantam2015cider}, BLEU~\cite{papineni2002bleu}, METEOR~\cite{banerjee2005meteor}, ROUGE~\cite{lin2004rouge}, and SPICE~\cite{anderson2016spice}. CIDEr-D is the most robust one among them. The higher these similarity metrics are, the more human-like the generated captions are.

2) We exploited CHAIRs and CHAIRi~\cite{rohrbach2018object} to quantitatively measure the bias degree of the generated captions to some objects. The lower CHAIRs and CHAIRi are, the less biased the captions are. Fig.~\ref{fig:intro2} gives some biased examples. Table~\ref{table:tab_kap} reports the results on the five similarity metrics and the two bias degree metrics, where the models are trained following the settings in Section~\ref{subsec:dataset}. 

3) To confirm whether each feature extraction module learns its module-specific knowledge, we calculated two metrics which are the recall of each module-specific word and the accuracy of the predicted module layout. Here we show how to calculate the recall of the nouns for \textsc{object} module. For the $i$-th image, we counted all the non-repetitive nouns from this images' ground-truth captions and denoted this number as $N_{gt}^i$. We also counted how many of these nouns non-repetitively appearing in the predicted caption as $N_{pre}^i$. Then the recall is $\sum_i{N_{pre}^i}/\sum_i{N_{gt}^i}$. We reported five different recalls in Table~\ref{tab:statistic}: nouns for \textsc{object} module; adjectives for \textsc{attribute} module; and verbs, prepositions, and quantifiers for \textsc{relation} module. To measure the module layout accuracy, for each generated word, we treated the maximum soft fusion weight $w$ (Eq.~\eqref{equ:soft_fuse}) as the predicted module and inspected whether it matches with the ground-truth module $w^*$. We report this accuracy in Table~\ref{tab:module_accuracy}.

\begin{figure}[t]
\centering
\includegraphics[width=1\linewidth,trim = 5mm 5mm 5mm 5mm,clip]{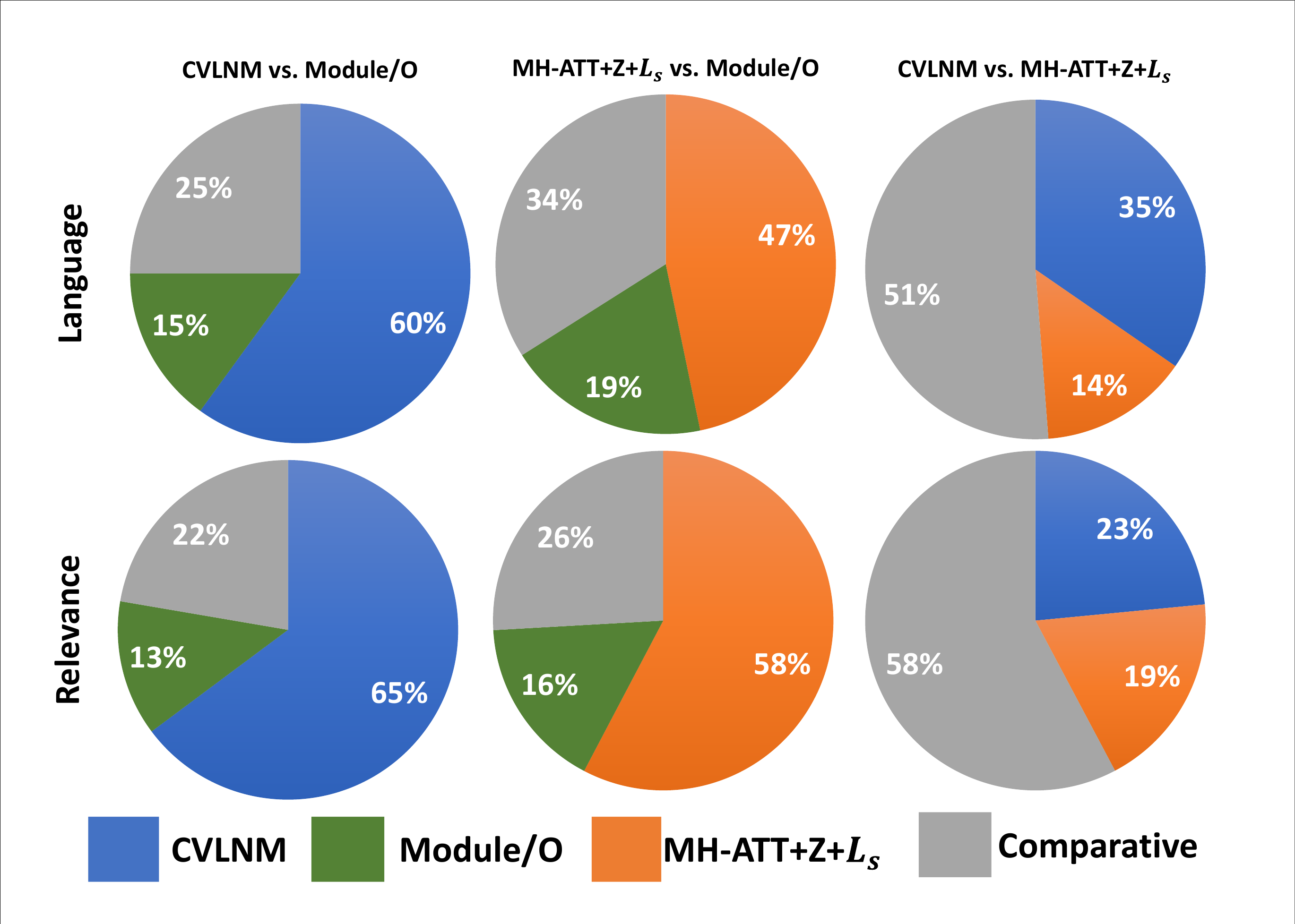}
  \caption{Each pie chart shows the comparison of two methods in human evaluation.}
\label{fig:he}
\end{figure}

4) We also invited 20 workers for human evaluation to test the qualities of the generated captions. We exhibited 50 images sampled from the test set to each worker and asked them to pairwise compare the captions generated from three models: Module/O, MH-ATT+$\mathcal{Z}$+$L_{s}$, and CVLNM. The captions were compared from two aspects: the language aspect that whether the generated captions are fluent and descriptive (the top three pie charts in Fig.~\ref{fig:he}); and the relevance aspect that whether the generated captions match with the images (the bottom three pie charts). 

\begin{table}[t]
\begin{center}
\caption{The performances of various baselines on Karpathy split. B@4, M, R, C, S, CHs, and CHi denote BLEU@4, METEOR, ROUGE-L, CIDEr-D, SPICE, CHAIRs, and CHAIRi, respectively. The symbols $\uparrow$ and $\downarrow$ mean the higher the better and the lower the better, respectively.}
\label{table:tab_kap}
\scalebox{0.78}{
\begin{tabular}{l c c c c c c c}
		\hline
		   Models  & B@4$\uparrow$ & M$\uparrow$ & R$\uparrow$ &  C$\uparrow$ & S$\uparrow$ & CHs$\downarrow$ & CHi$\downarrow$ \\
          \hline 
          Module/O & $37.5$ & $27.7$ & $57.5$ & $123.1$ & $21.0$ & $13.8$ & $9.4$\\
          Module/A & $37.3$ & $27.4$ & $57.1$ & $121.9$ & $20.9$ & $14.3$ & $9.8$\\
          Module/R & $37.9$ & $27.8$ & $57.8$ & $123.8$ & $21.2$ & $13.6$ & $9.3$\\
          Col/1 & $38.2$ & $27.9$ & $58.1$ & $125.6$ & $ 21.2$ & $13.5$ & $9.1$\\
          Col/H & $38.4$ & $28.2$ & $58.3$ & $126.1$ & $ 21.3$ & $11.7$ & $8.2$\\
          Col/S (MH-ATT+$\mathcal{Z}$) & $38.5$ & $28.3$ & $58.4$ & $127.3$ & $21.6$ & $11.3$ & $7.8$\\
          LSTM & $38.2$ & $28.0$ & $58.2$ & $125.7$ & $21.2$ & $12.4$ & $8.5$ \\
          LSTM+$\mathcal{Z}$ & $38.4$ & $28.1$ & $58.3$ & $126.3$ & $21.4$ & $12.3$ & $8.4$ \\
          MH-ATT & $38.3$ & $28.0$ & $58.4$ & $126.7$ & $21.2$ & $11.9$ & $8.2$  \\
          MH-ATT+$\mathcal{Z}$+Reason & $38.7$ & $28.3$ & $58.6$ & $128.6$ & $21.7$ & $11.0$ & $7.6$\\
          MH-ATT+$\mathcal{Z}$+$L_{s}$ & $38.9$ & $28.4$ & $58.5$ & $128.4$ & $21.9$ & $10.8$ & $7.4$\\
          CVLNM  & $\bm{39.4}$ & $\bm{28.7}$ & $\bm{59.1}$ & $\bm{129.5}$ & $\bm{22.2}$ & $\bm{10.5}$ & $\bm{7.0}$ \\
          \hline 
          CVLNM ($\lambda=0.1$)  & 39.1 & 28.5 & 59.0 & 129.2 & 22.0 & 10.6 & 7.1 \\
          CVLNM ($\lambda=1$) & $39.4$ & $28.8$ & $59.0$ & $129.4$ & $22.1$ & $10.5$ & $6.9$ \\
          CVLNM ($\lambda=5$) & $39.1$ & $28.4$ & $58.8$ & $129.2$ & $21.8$ & $10.7$ & $7.1$ \\
          CVLNM+  & $39.7$ & $28.9$ & $59.4$ & $130.1$ & $22.4$ & $10.5$ & $6.9$ \\ \hline
\end{tabular}
}
\end{center}
\end{table}

\begin{table}[t]
\begin{center}
\caption{The recalls (\%) of five part-of-speech words, where nouns correspond to \textsc{object} module; adjectives correspond to \textsc{attribute} module; and verbs, prepositions, and quantifiers correspond to \textsc{relation} module.}
\label{tab:statistic}
\scalebox{0.85}{
\begin{tabular}{l c c c c c}
		\hline
		   Models   & nouns & adjectives & verbs & prepositions & quantifiers \\ \hline
           Module/A & $42.4$ & $12.4$ & $20.2$ & $41.7$ & $14.3$ \\
           Module/O & $44.5$ & $11.5$ & $21.8$ & $42.6$ & $17.1$ \\ 
           Module/R & $44.3$ & $11.3$ & $22.8$ & $43.5$ & $22.3$ \\
           LSTM & $45.2$ & $13.1$ & $23.1$ & $43.6$ & $23.4$ \\
           MH-ATT & $45.5$ & $13.5$ & $23.4$ & $43.9$ & $24.2$ \\
           MH-ATT+$\mathcal{Z}$ & $46.3$ & $14.8$ & $23.8$ & $44.2$ & $26.8$ \\
           MH-ATT+$\mathcal{Z}$+$L_s$ & $\bm{47.6}$ & $\bm{16.5}$ & $\bm{24.4}$ & $\bm{45.1}$ & $\bm{29.6}$ \\ \hline
\end{tabular}
}
\end{center}
\end{table}

\begin{table}[t]
\begin{center}
\caption{The ratios (\%) of different words among all the generated words when cutting off a module, \eg, ``-Module/O'' means cutting off \textsc{object} module.}
\label{table:womodule}
\scalebox{0.9}{
\begin{tabular}{l c c c c }
		\hline
		   Module  & \textsc{object} & \textsc{attribute} & \textsc{relation} & \textsc{function}  \\
          \hline 
          all modules   & 54 & 10  & 27 & 9  \\
          -Module/O     & 24 & 31 & 34 & 11  \\
          -Module/A     & 63 & 1  & 27 & 9   \\
          -Module/R     & 62 & 10  & 17 & 11 \\
          -Module/F     & 61 & 8  & 30 & 1   \\
          \hline
\end{tabular}
}
\end{center}
\end{table}

\begin{table}[t]
\begin{center}
\caption{The accuracy (\%) of the predicted module layout by the module controller. The column ``Average'' means the average accuracy of all four modules.}
\label{tab:module_accuracy}
\scalebox{0.80}{
\begin{tabular}{l c c c c c}
		\hline
		   Models   & \textsc{object} & \textsc{attribute} & \textsc{relation} & \textsc{function} & Average \\ \hline
           LSTM & $56.3$ & $39.7$ & $75.3$ & $27.7$ & $59.4$  \\
           MH-ATT & $62.2$ & $42.3$ & $78.2$ & $29.1$ & $64.2$ \\
           MH-ATT+$\mathcal{Z}$ & $66.2$ & $46.8$ & $83.8$ & $32.5$ & $68.6$ \\
           MH-ATT+$\mathcal{Z}$+$L_s$ & $\bm{74.9}$ & $\bm{55.7}$ & $\bm{93.0}$ & $\bm{39.0}$ & $\bm{77.6}$\\ \hline
\end{tabular}
}
\end{center}
\end{table}

\begin{figure*}[t]
\centering
\includegraphics[width=1\linewidth,trim = 5mm 5mm 5mm 5mm,clip]{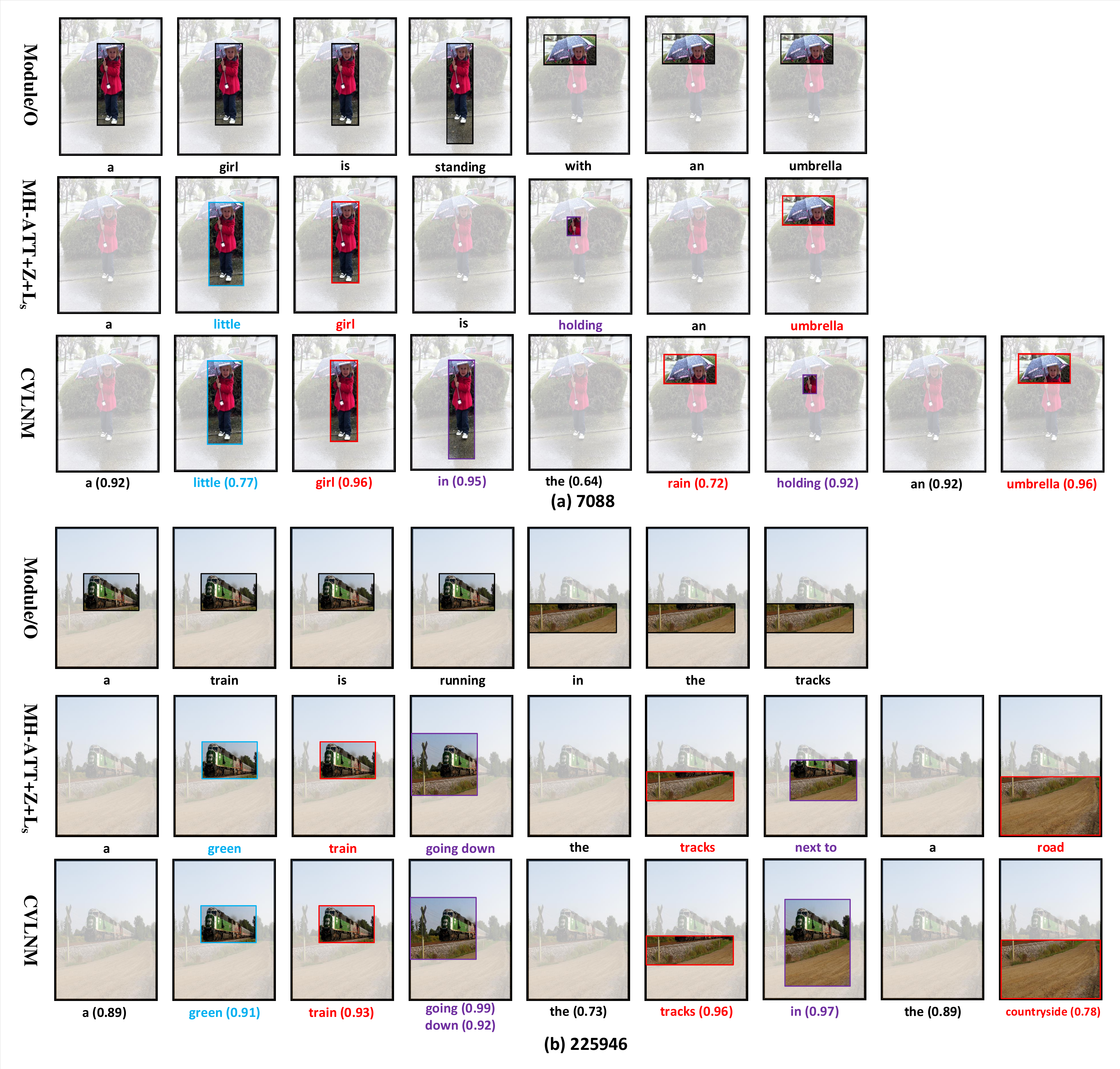}
  \caption{Visualizations of the captioning processes Module/O, MH-ATT+$\mathcal{Z}$+$L_s$, and CVLNM. Different colours refer to different modules, \ie, red/blue/purple/black for \textsc{object}/\textsc{attribute}/\textsc{relation}/\textsc{function} module. In Module/O, we use black boxes to show the attended regions. In the other two methods, for simplicity, we only show the module with the largest soft module collocation weight (Eq.~\eqref{equ:soft_fuse}) and the corresponding region with the largest attention weight (Eq.~\eqref{equ:attend}). In CVLNM, the number in the bracket is the corresponding soft module weight.
  }
\label{fig:demo}
\end{figure*}
\subsubsection{Empirical Answers}
\label{subsec:empirical_answers}
\noindent\textbf{A1:}
From Table~\ref{tab:statistic}, we find that each module prefers to generate more corresponding module-specific words, \eg, among three single-module models, Module/O has the highest recall of nouns. Moreover, in Table~\ref{tab:module_accuracy}, we find that even when a simple module controller (the baseline LSTM) is applied to predict the module layout, the accuracy is still much better than randomly choosing (which gives rise to 25\% average accuracy) \footnote{Intuitively, by randomly choosing, 
at each time, the probability of choosing the right one from 4 modules is $1/4=25\%$.}. Both observations validate that \textit{by using diverse inductive biases to design four modules, they can learn module-specific knowledge to generate more corresponding words and to predict better module layouts.}

From Table~\ref{table:womodule}, we can find that when one module is cut off, the ratio of the corresponding word will largely decrease, \eg, after cutting off \textsc{attribute} module, \textsc{attribute} related words will decrease from 10\% to 1\%. However, these words will not be eliminated. The major reason is that the other modules and the language decoder (two LSTM layers in Eq.~\eqref{equ:top_down}) will leak certain attribute related information for generating attributes.

\noindent\textbf{A2:} 
As shown in Table~\ref{table:tab_kap}, when we fuse the modules by three different strategies, the performances are all improved compared with the single-module baselines, \eg, Col/1 is better than Module/R, which answers the first part of Q2 that \textit{fusing modules will generate better captions}.

To answer the second part of Q2, we first observe that in Table~\ref{table:tab_kap}, both the hard selection (Col/H) and the soft fusion (Col/S) achieve better performances than equally exploiting the modules (Col/1), which suggests that \textit{the modules should be discriminatively fused}. Furthermore, we find that Col/S outperforms Col/H. One possible reason is that the module layout is parsed from the partially generated caption, which is not perfect enough and the complement of the other modules is necessary.

\noindent\textbf{A3:}
In Table~\ref{table:tab_kap},~\ref{tab:statistic} and~\ref{tab:module_accuracy}, we can find that when MH-ATT is applied (MH-ATT vs. LSTM) and when $\mathcal{Z}$ is used (MH-ATT+$\mathcal{Z}$ vs. MH-ATT), the similarity scores, the recalls, and the layout accuracies are all improved. All of these observations confirm that compared with the module controller in our preliminary work~\cite{yang2019learning}, which corresponds to the baseline LSTM, \textit{applying MH-ATT and inputting $\mathcal{Z}$ will enhance the effectiveness of the module controller, which then gives rise to more distinguishable modules and better captions}.


\noindent\textbf{A4:} 
By comparing MH-ATT+$\mathcal{Z}$+$L_s$ with MH-ATT+$\mathcal{Z}$ in Tables~\ref{table:tab_kap},~\ref{tab:statistic}, and~\ref{tab:module_accuracy}, we find that MH-ATT+$\mathcal{Z}$+$L_s$ achieves better results. Specifically, the improvements of the word recall and the module accuracy validate that \textit{the syntax loss encourages each module to learn more distinguishable knowledge}, and higher CIDEr scores and lower bias metrics suggest that \textit{more distinguishable modules lead to more human-like and less biased captions}.

\noindent\textbf{A5:} 
In Table~\ref{table:tab_kap} and Fig~\ref{fig:he}, we respectively observe that when \textsc{reason} module is applied, the similarity scores are improved (MH-ATT+$\mathcal{Z}$+\textsc{reason} vs. MH-ATT+$\mathcal{Z}$) and the captions are considered better by humans (CVLNM vs. MH-ATT+$\mathcal{Z}$+$L_s$), especially in terms of language descriptiveness. Both comparisons suggest that \textsc{reason} module improves the quality of the generated captions by approximating human-like commonsense reasoning. Also, as shown in Figure~\ref{fig:demo} (a), we observe that in CVLNM, the word ``rain'' is generated by the region covering ``umbrella'' and ``wet floor'', while when \textsc{reason} module is not used, the word ``rain'' is not generated in MH-ATT+$\mathcal{Z}$+$L_s$. 

\noindent\textbf{A6:} 
In Tables~\ref{table:tab_kap},~\ref{tab:statistic},~\ref{tab:module_accuracy} and Fig.~\ref{fig:he}, we find that by sequentially incorporating the module controller, the syntax loss, and \textsc{reason} module into the pipeline, more distinguishable modules can be learned, lower bias degree can be achieved, and better captions can be generated. Fig.~\ref{fig:demo} gives two examples to show how CVLNM collocates the modules to generate the captions, which also demonstrates that our CVLNM has better interpretability than the single module based captioner.

\begin{figure}[t]
\centering
\includegraphics[width=1\linewidth,trim = 5mm 5mm 5mm 5mm,clip]{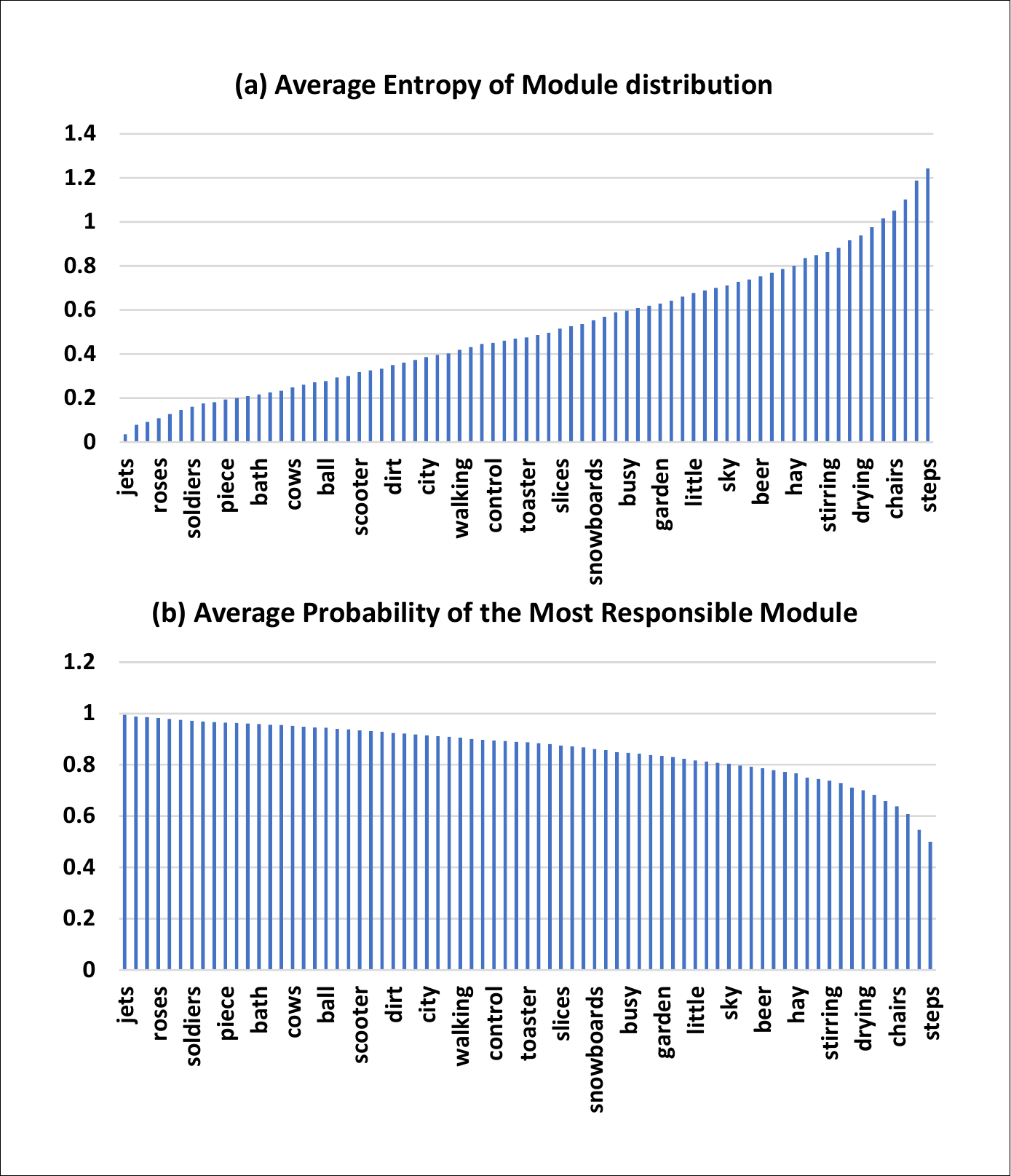}
  \caption{The average entropy of the module distribution and the average probability of the most responsible module for each word.}
\label{fig:module_distribution_few}
\end{figure}
\noindent\textbf{A7:} 
Figure~\ref{fig:module_distribution_few} shows the average entropy and the average probability for each word. It can be found that for most words, the entropy is small and the probability is large, \eg, a large number of words have a probability larger than 0.8, which indicates that the module distributions are sharp in most cases. Figure~\ref{fig:demo} also gives two examples, where we can see that most words are generated mainly from one module.

\noindent\textbf{A8:} 
The bottom part of Table~\ref{table:tab_kap} shows the results of using different hyperparameters. It can be observed that using different $\lambda$ does not largely affect the performances, \eg, when setting $\lambda=5$, CIDEr-D only changes 0.3 compared with the original CVLNM. Also, by using more object and attribute annotations to pre-train the CNN for extracting features, the performances have certain improvements, \eg, the CIDEr-D increases from 129.5 (CVLNM) to 130.1 (CVLNM+).

\begin{table*}
\begin{center}
\caption{The performances of various methods on MS-COCO Karpathy split trained by cross-entropy loss only (left part) and the mixture of cross-entropy loss and self-critical reward (right part).}
\label{table:tab_xe}
\scalebox{1}{
\begin{tabular}{l | c c c c c | c c c c c}
		\hline
		Loss  & \multicolumn{5}{c|}{Cross-Entropy}  & \multicolumn{5}{c}{Cross-Entropy \& Self-Critical} \\ \hline
		   Models   & B@4 & M & R &  C & S & B@4 & M & R &  C & S\\ \hline
           SCST~\cite{rennie2017self}       & $30.0$ & $25.9$ & $53.4$ & $99.4$ & $-$ & $34.2$ & $26.7$ & $55.7$ & $114.0$ & $-$\\ 
           StackCap~\cite{gu2017stack}    & $35.2$ & $26.5$ & $-$ & $109.1$ & $-$ & $36.1$ & $27.4$ & $-$ & $120.4$ & $-$ \\ 
           NBT~\cite{lu2018neural}   & $34.7$ & $27.1$ & $-$ & $108.9$ & $20.1$ & $-$ & $-$ & $-$ & $-$ & $-$    \\
           Up-Down~\cite{anderson2018bottom}  & $36.2$ & $27.0$ & $56.4$ & $113.5$ & $20.3$ & $36.3$ & $27.7$ & $56.9$ & $120.1$ & $21.4$ \\ 
           RFNet~\cite{jiang2018recurrent}   & $37.0$ & $27.9$ & $57.3$ & $116.3$ & $20.8$ & $37.9$ & $28.3$ & $58.3$ & $125.7$ & $21.7$  \\ 
           GCN-LSTM~\cite{yao2018exploring}   & $36.8$ & $27.9$ & $57.0$ & $116.3$ & $20.9$ & $38.2$ & $28.5$ & $58.3$ & $127.6$ & $22.0$  \\ 
           CAVP~\cite{zha2019context} & $-$ & $-$ & $-$ & $-$ & $-$  & $38.6$ & $28.3$ & $58.5$ & $126.3$ & $21.6$ \\
           LBPF~\cite{qin2019look}    & $37.4$ & $28.1$ & $57.5$ & $116.4$ & $21.2$ & $38.3$ & $28.5$ & $58.4$ & $127.6$ & $22.0$\\
           SGAE~\cite{yang2019auto}    & $36.9$ & $27.7$ & $57.2$ & $116.7$ & $20.9$ & $38.4$ & $28.4$ & $58.6$ & $127.8$ & $22.1$ \\
           CNM~\cite{yang2019learning}  & $37.1$ & $27.9$ & $57.3$ & $116.6$ & $20.8$ & $38.9$ & $28.4$ & $58.8$ & $127.9$ & $22.0$\\
           \re{HIP+Up-Down~\cite{yao2019hierarchy}}  & \re{37.0} & \re{28.1} & \re{57.1} & \re{116.6} & \re{21.2} & \re{38.2} & \re{28.4} & \re{58.3} & \re{127.2} & \re{21.9} \\
           \re{HIP+GCN-LSTM~\cite{yao2019hierarchy}}  & \re{$\bm{38.0}$} & \re{$\bm{28.6}$} & \re{$\bm{57.8}$} & \re{$\bm{120.3}$}&  \re{$\bm{21.4}$} & \re{$39.1$} & \re{$\bm{28.9}$}& \re{59.2}& \re{$\bm{130.6}$}& \re{$22.3$} \\
           \re{ETA~\cite{li2019entangled}}  & \re{37.1}& \re{28.2}& \re{57.1}& \re{117.9} & \re{21.4} & \re{39.3} & \re{28.8} & \re{58.9} & \re{126.6} & \re{22.7} \\ 
           \re{ToW~\cite{herdade2019image}} & \re{35.5}& \re{28.0}& \re{56.6}& \re{115.3} & \re{21.2} & \re{38.6} & \re{28.7} & \re{58.4} & \re{128.3} & \re{22.6} \\
           CVLNM  & $37.3$ & $28.2$ & $57.6$ & $117.1$ & $21.2$ & $39.4$ & $28.7$ & $59.1$ & $129.5$ & $22.2$\\ 
           AoANet~\cite{huang2019attention}  & $37.2$ & $28.4$ & $57.5$ & $119.8$ & $21.3$ & $38.9$ & $29.2$ & $58.8$ & $129.8$ & $22.4$ \\
           CVLNM*  & $37.4$ & $28.3$ & $\bm{57.8}$ & $119.2$ & $21.3$ & $\bm{39.6}$ & $29.1$ & $\bm{59.4}$ & $130.2$ & $\bm{22.5}$\\ 
           \hline 
\end{tabular}
}
\end{center}
\end{table*}
\begin{table*}
\begin{center}
\caption{The performances of various compared image captioners on the online MS-COCO test server.}
\label{table:tab_online}
\scalebox{1}{
\begin{tabular}{l c c c c c c c c c c}
		\hline
		 Model  & \multicolumn{2}{c}{B@3} & \multicolumn{2}{c}{B@4} &\multicolumn{2}{c}{M} &\multicolumn{2}{c}{R-L} & \multicolumn{2}{c}{C-D}\\ \hline 
		 Metric  & c5 & c40 & c5 & c40 & c5 & c40 & c5 & c40 & c5 & c40 \\ \hline
           SCST~\cite{rennie2017self}     & $47.0$ & $75.9$ & $35.2$ & $64.5$ & $27.0$ & $35.5$ & $56.3$ & $70.7$ & $114.7$ & $116.0$  \\
           LSTM-A~\cite{yao2017boosting}   & $47.6$ & $76.5$ & $35.6$ & $65.2$ & $27.0$ & $35.4$ & $56.4$ & $70.5$ & $116.0$ & $118.0$  \\ 
           StackCap~\cite{gu2017stack}     & $46.8$ & $76.0$ & $34.9$ & $64.6$ & $27.0$ & $35.6$ & $56.2$ & $70.6$ & $114.8$ & $118.3$   \\ 
           Up-Down~\cite{anderson2018bottom}    & $49.1$ & $79.4$ & $36.9$ & $68.5$ & $27.6$ & $36.7$ & $57.1$ & $72.4$ & $117.9$& $120.5$  \\ 
           RFNet~\cite{jiang2018recurrent}      & $50.1$ & $80.4$ & $38.0$ & $69.2$ & $28.2$ & $37.2$ & $58.2$ & $73.1$ & $122.9$& $125.1$    \\
           CAVP~\cite{liu2018context}            & $50.0$ & $79.7$ & $37.9$ & $69.0$ & $28.1$ & $37.0$ & $58.2$ & $73.1$ & $121.6$& $123.8$    \\
           SGAE~\cite{yang2019auto}     & $50.1$   & $79.6$ & $37.8$ & $68.7$& $28.1$ & $37.0$ & $58.2$ & $73.1$ & $122.7$ & $125.5$\\ 
           CNM~\cite{yang2019learning}    & $50.2$ & $79.8$ & $38.4$ & $69.3$& $28.2$ & $37.2$ & $58.4$ & $73.4$ & $123.8$ & $126.0$\\
           \re{AoANet~\cite{huang2019attention}} & \re{$51.4$} & \re{$81.3$} & \re{$39.4$} & \re{$\bm{71.2}$} & \re{$29.1$} & \re{$\bm{38.5}$} & \re{$58.9$} & \re{$74.5$} & \re{$126.9$} & \re{$129.6$}\\
           \re{HIP+GCN-LSTM~\cite{yao2019hierarchy}}  &
           $\re{\bm{51.5}}$ & $\re{\bm{81.6}}$ & $\re{\bm{39.3}}$ & $\re{\bm{71.0}}$ & $\re{\bm{28.8}}$ & $\re{\bm{38.1}}$ & $\re{\bm{59.0}}$ & $\re{\bm{74.1}}$ & $\re{\bm{127.9}}$ & $\re{\bm{130.2}}$ \\
           \re{ETA~\cite{li2019entangled}} & \re{50.9} & \re{80.4} & \re{38.9} & \re{70.2} & \re{28.6} & \re{38.0} & \re{58.6} & \re{73.9} & \re{122.1} & \re{124.4} \\      
           CVLNM    & $50.5$ & $80.4$& $38.5$ & $70.0$& $28.7$ & $38.0$ & $58.5$ & $73.7$ & $125.2$ & $127.8$\\ 
           CVLNM-Ensemble    & $51.2$ & $81.2$& $39.1$ & $70.9$& $28.8$ & $38.2$ & $58.8$ & $74.2$ & $126.6$ & $129.1$\\ 
           \hline
\end{tabular}
}
\end{center}
\end{table*}
\begin{table}
\begin{center}
\caption{The CIDEr-D deterioration \re{(the difference of scores)} of using fewer training sentences. The values in the bracket are the real CIDEr-D scores.}
\label{table:tab_fewer}
\scalebox{0.7}{
\begin{tabular}{l c c c c c}
		\hline
		   X   & 5 & 4 & 3 & 2 & 1 \\ \hline
          Module-O\&X & $0 (123.1)$ & $0.9 (122.2)$ & $2.3 (120.8)$ & $4.1 (119.0)$ & $6.8 (116.3)$ \\ 
          MH-ATT+$\mathcal{Z}$\&X & $0 (127.1)$ & $0.5 (126.6)$ & $1.4 (125.7)$ & $3.0 (124.1)$ & $4.5 (122.6)$ \\
          MH-ATT+$\mathcal{Z}$+$L_s$\&X & $0 (128.4) $ & $\bm{0.3} (128.1)$ & $\bm{1.0}(127.4)$ & $\bm{2.1}(126.3)$ & $\bm{3.6}(124.8)$ \\
          \hline
\end{tabular}
}
\end{center}
\end{table}

\subsection{Comparisons with the State-of-The-Art Models}\label{sec:sota}
\noindent\textbf{Comparing Methods.} We compared CVLNM with the following state-of-the-art image captioners: \textbf{SCST}~\cite{rennie2017self}, \textbf{Up-Down}~\cite{anderson2018bottom}, \textbf{NBT}~\cite{lu2018neural}, \textbf{CAVP}~\cite{zha2019context}, \textbf{RFNet}~\cite{jiang2018recurrent}, \textbf{LBPF}~\cite{qin2019look}, \textbf{GCN-LSTM}~\cite{yao2018exploring}, \textbf{SGAE}~\cite{yang2019auto}, \re{\textbf{HIP}~\cite{yao2019hierarchy}, \textbf{ETA}~\cite{li2019entangled}, \textbf{ToW}~\cite{herdade2019image} and \textbf{AoANet}~\cite{huang2019attention}}. Some of them can be treated as simple versions of our CVLNM, \eg, NBT only uses \textsc{object} module, CAVP and GCN-LSTM apply different \textsc{relation} modules, and Up-Down integrates \textsc{object} and \textsc{attribute} modules. RFNet uses various CNN architectures to extract visual features, which owns a wider encoder compared with some other methods, and its comparison with our CVLNM is to prove that only a wider encoder is not enough for better captions. SGAE not only exploits the additional scene graph annotations but also uses Graph Neural Network (GNN) to embed such visual scene graphs to transfer linguistic inductive bias from the pure language domain to the vision-language domain for more descriptive captions. In CVLNM, \textsc{reason} module has a similar function which transfers linguistic inductive bias.
Moreover, we compared our CVLNM with our previous CNM~\cite{yang2019learning} to show that by a more advanced module controller and \textsc{reason} module, better captions can be generated. 

For AoANet, it uses four stronger techniques compared with our CVLNM. First, their Attention on Attention (AoA) is indeed a more effective attention mechanism than the one (Eq.~\eqref{equ:attend}) used in our CVLNM, which comes from the classic Top-Down attention~\cite{anderson2018bottom}. Second, AoANet stacks 6 AoA layers as their encoder, which is more similar to the architecture of Transformer~\cite{vaswani2017attention} instead of the classic Top-Down RNN. Third, AoANet uses a better strategy to adjust the learning rate. In particular, they anneal the learning rate by 0.5 when the CIDEr-D score on the validation split does not improve. Fourth, AoANet uses scheduled sampling~\cite{bengio2015scheduled} when the cross-entropy loss is used to train their network, which can largely improve the CIDEr-D score. Since the latter two general techniques can be used in any captioner, we also applied them to train a model named CVLNM*. Specifically, when the CIDEr-D score on the validation split did not improve for 3 epochs, we annealed the learning rate by 0.5. We also increased the scheduled sampling probability from 0 at the beginning by 0.05 every 5 epochs as AoANet when the cross-entropy loss was used. 

\noindent\textbf{Results and Analysis.}
The left and right parts of Table~\ref{table:tab_xe} report the performances of various methods trained by cross-entropy loss and the mixture of cross-entropy loss and self-critical reward, respectively. From both parts, we can see that CVLNM outperforms the other Top-Down RNN based models and it achieves a new state-of-the-art 129.5 CIDEr-D score when the mixture training losses are applied. Specifically, by using four distinguishable feature extraction modules, an advanced module controller, the syntax loss, and \textsc{reason} module, our CVLNM significantly outperforms the single-module models, \eg, NBT, CAVP, and GCN-LSTM; the model with a wider encoder, \eg, RFNet; and the GNN based model, \eg, SGAE. Due to the more advanced module controller and \textsc{reason} module, CVLNM outperforms CNM. Compared with AoANet, after using the two advanced training strategies as them, our CVLNM* can achieve comparable performances. When only the cross-entropy loss is used, AoANet's CIDEr-D is slightly better; when the mixture losses are used, our CVLNM* obtains a slightly higher CIDEr-D score. 
\re{Since HIP~\cite{yao2019hierarchy} applies more fine-grained segmentation annotations, the extracted visual features are better than our features trained by detection. However, we still achieve better performances than HIP+Up-Down and comparable performances with HIP+GCN-LSTM (which uses GCN~\cite{yao2018exploring} as the encoder to further improve the captioning model). Both ETA~\cite{li2019entangled} and ToW~\cite{herdade2019image} use Transformer as the backbone while our CVLNM uses more traditional LSTM-based architecture and still outperforms them. Such comparisons also confirm the validness of our model.
}

\re{Moreover, we submitted our CVLNM to the online server and report the scores in Table~\ref{table:tab_online}, where CVLNM denotes the single-model and CVLNM-Ensemble denotes the ensembled models (We follow the previous researchers~\cite{anderson2018bottom,huang2019attention,jiang2018recurrent} to ensemble 4 models.). In the online testing, HIP+GCN-LSTM utilizes SENet-154 as the backbone of Faster R-CNN and Mask R-CNN (while we use ResNet as the backbone of Faster R-CNN and does not use Mask R-CNN) to extract the visual features and thus achieves the best performances. For AoANet, which uses four stronger strategies to train their models, our CVLNM-Ensemble can still achieve comparable results. When compared with the other models like ETA~\cite{li2019entangled} and ToW~\cite{herdade2019image}, our models can outperform them.}

\subsection{Few-Shot Image Captioning}
\label{sec:few_shot}
\noindent\textbf{Comparing Methods.}
In this setting, for each image, we randomly sampled $X$ captions from 5 ground-truth captions to train our models. To avoid confusion with the methods in Section~\ref{sec:experiment}, we add \textbf{\&X} to each method to denote that only $X$ captions were provided, \eg, \textbf{Module-O\&2} means 2 captions of each image were sampled to train Module-O. We compared the following baselines to see how each component affects the robustness: 1) \textbf{Module-O\&X} to test only one module, 2) \textbf{MH-ATT+$\mathcal{Z}$\&X} to test the soft fusion, and 3) \textbf{MH-ATT+$\mathcal{Z}$+$L_s$\&X} to test the syntax loss. Noteworthy, for fair comparisons with Module-O\&X, we did not pre-train \textsc{attribute} module with attribute labels and did not use \textsc{reason} module in MH-ATT+$\mathcal{Z}$\&X and MH-ATT+$\mathcal{Z}$+$L_s$\&X. The results are reported in Table~\ref{table:tab_fewer}, where the values in the table are the deterioration of CIDEr-D scores compared with the model trained by all 5 sentences, whose scores are given in the bracket.

\noindent\textbf{Results and Analysis.}
From Table~\ref{table:tab_fewer}, we can find that all the models achieve lower performances when fewer training sentences are provided. Interestingly, we observe that our MH-ATT+$\mathcal{Z}$+$L_s$\&X can almost halve the CIDEr-D deterioration compared to Module/O\&X. For example, when only 1 caption is available, the deterioration of MH-ATT+$\mathcal{Z}$+$L_s$ is 3.6, while Module-O is 6.8. Also, it can be found that when we apply the soft fusion strategy and impose the syntax loss to further regularize the learning, the deterioration caused by fewer training samples becomes less severe. Such observations suggest that \textit{by decomposing the captioning into a series of sub-tasks solved by distinguishable modules and by applying suitable fusion and learning strategies, the whole system becomes more robust}.

\section{Conclusions}
We proposed to follow the principle of modular design for image captioning. In particular, we presented a novel module network: learning to Collocate Visual-Linguistic Neural Modules (CVLNM), which can generate captions by filling the contents into the collocated patterns. In CVLNM, we designed four modules with diverse inductive biases in the encoder to extract features and one \textsc{reason} module in the decoder to approximate commonsense reasoning. A multi-head self-attention based module controller was designed to dynamically collocate the feature extraction modules since only the partially observable caption is available. A syntax based loss was imposed on the module controller to further guarantee each feature extraction module to learn module-specific knowledge as well as encouraging the controller to learn human-like sentence patterns. We validated the effectiveness and robustness of our CVLNM by extensive ablations and comparisons with the state-of-the-art models on MS-COCO. The experiment results substantially demonstrated that our CVLNM not only generates more human-like captions but also suffers less deterioration when fewer captions are provided for training. 

\section*{Acknowledgments}
This work is supported by Singapore MOE AcRF Tier 2 MOE2019-T2-2-062.

\bibliographystyle{spbasic}      
\bibliography{egbib.bib}



\end{document}